\documentclass[conf]{new-aiaa}

\usepackage[utf8]{inputenc}
\usepackage{graphicx}
\usepackage{amsmath}
\usepackage[version=4]{mhchem}
\usepackage{siunitx}
\usepackage{longtable,tabularx}
\usepackage{subfiles}
\usepackage{subcaption}

\usepackage{bm}
\usepackage{mwe}
\usepackage{algorithm}
\usepackage{algpseudocode}
\usepackage{float}
\title{Three Dimensional Hydrodynamic Flow-Based Collision Avoidance for UAV Formations Facing Emergent Dynamic Obstacles}

\author{Suguru Sato\footnote{Ph.D. student, Mechanical and Aerospace Engineering, 701 S. Nedderman Dr., Arlington, TX, 76019, AIAA student member.} and Kamesh Subbarao \footnote{Professor, Mechanical and Aerospace Engineering, 701 S. Nedderman Dr., Arlington, TX, 76019, AIAA Associate Fellow.}}
\affil{The University of Texas at Arlington, Arlington, Texas, 76019}

\begin{document}

\maketitle

\begin{abstract}

This paper presents a three-dimensional, hydrodynamics-inspired collision avoidance framework for uncrewed aerial vehicle (UAV) formations operating in dynamic environments. When moving obstacles enter a UAV's sensing region, they are modeled as three dimensional doublets or ellipsoids that generate local velocity fields, guiding nearby UAVs to execute smooth, collision-free maneuvers without trajectory discontinuities or explicit trajectory replanning. This flow-based approach enables real-time operation and interpretable behavior by leveraging the nature of fluid flow around obstacles via the harmonic properties of Laplace's equation, inherently avoiding local minima common in traditional potential field methods. To establish and maintain coordination among the UAVs, a Virtual Rigid Body (VRB) formation strategy is integrated, ensuring that formation geometry and trajectory tracking are preserved. Simulation results demonstrate the feasibility and scalability of the method for both individual and multi-UAV scenarios with multiple formation geometries encountering moving obstacles. The proposed approach achieves safe, smooth, and computationally efficient avoidance maneuvers suitable for real-time and practical applications.

\end{abstract}

\section{Introduction}

Uncrewed Aerial Vehicles (UAVs) have become essential in both civilian and military applications due to their versatility, low operational cost, and ability to operates within environments that are unsafe or unreachable for humans. Moreover, UAV formation flight, wherein multiple UAVs operates within close proximity while maintaining precise relative and coordinated movement \cite{sato_2024}, offers significant benefits in missions such as reconnaissance, mapping, search-and-rescue, and package delivery with enhanced scalability, sensing redundancy, and energy efficiency. However, as more and more UAVs and UAV formations are deployed due to their beneficial capabilities, the environments increasingly become cluttered and dynamic. Therefore, collision avoidance, particularly against emergent moving obstacles, becomes critical challenge for mission safety and success.

Existing collision avoidance methods can be broadly classified into three methods: reactive, predictive, and learning-based approach. Reactive methods, such as the potential field \cite{khatib_1985} and the dynamic window approach \cite{fox_1997} generate avoidance behaviors directly from sensor data to allow fast and intuitive responses. While these approaches are simple and interpretable, they are more likely to suffer from oscillation behavior, jittering motion, and local minima, which make them unreliable in dense environments. Predictive methods, including Model Predictive Control (MPC)-based approaches \cite{richards_2024} and Velocity Obstacle frameworks \cite{fiorini_1998}, predicts future collision by estimating both vehicle and obstacle motion over the specified time horizon. Although these methods are effective, their reliance on accurate prediction and relatively high computational demands limit real-time and real-world execution and applicability. Learning-based approaches, such as reinforcement learning \cite{tai_2017}, have recently shown promissing results in achieving adaptive avoidance policies, but challenges remain in interpretability, safety certification, and generalization across environments.

Alongside these approaches, researchers have explored hydrodynamics-inspired methods, leveraging behavior of fluid flow around obstacles to generate smooth, efficient, and interpretable trajectories. Akishita et al. \cite{akishita_1990} introduced the concept of modeling obstacles as two-dimensional doublets within potential flow fields governed by Laplace's equation, which enables mobile robots to follow streamlines that naturally and smoothly avoid collisions. Waydo and Murray \cite{waydo_2003} and Szulcz\'nski et al. \cite{Szulczynski_2011, Szulczynski_2012} extended these concepts using harmonic and stream functions to ensure collision-free motion around moving obstacles modeled as doublets and ellipses. Palm and Driankov \cite{palm_2014} later combined velocity and force potentials to handle multi-agent interactions. Advanced extensions such as in Liu et al. \cite{liu_2016} proposed a 3D collision avoidance by interfered fluid dynamical system for efficient path planning and smooth UAV maneuver. Another extension by Liu et al \cite{liu_2021} proposed a path planning method incorporating virtual fluid field and CFD. Cai et al. \cite{cai_2021}, introduced a 3D stream-function-based obstacle avoidance for Autonomous Underwater Vehicles (AUVs) by extending a 2D fluid mechanics approach to a 3D environment. Recently, Zhang et al. \cite{zhang_2025} proposed the Formation Interfered Fluid Dynamical System (FIFDS), enabling cooperative UAV formation navigation through parallel streamline tracking in dense environments.

While these hydrodynamic flow-based approaches offer smooth and interpretable behaviors coming from behavior of fluid flow, most prior work was limited to 2D motion or single-vehicle scenarios, without consideration of maintaining predefined 3D formations or trajectory tracking in dynamic environments. Furthermore, integration of hydrodynamic flow-based flow modeling with formation-keeping strategies, which is essential for coordinated UAV operations, has not been fully addressed.

To overcome the challenges in the existing collision avoidance methods and the gap in the hydrodynamic flow-based approaches, this study introduces a three-dimensional hydrodynamic flow-based collision avoidance framework for UAV formations operating along predefined trajectories. In the proposed approach, emergent obstacles detected within a UAV's sensing range are modeled as 3D doublets or ellipsoids, which generate local velocity field resembling incompressible and irrotational fluid flow around a solid body. UAVs follow this induced flow to perform smooth, collision-free maneuvers without explicit trajectory replanning or reliance on future obstacle state prediction. To establish and maintain formation of UAVs and ensure the trajectory tracking fidelity of the formation, a Virtual Rigid Body (VRB) formation strategy is incorporated.

By combining the responsiveness of reactive control with the physics-based flow modeling, the proposed method provides a lightweight, interpretable, and scalable solution with real-time capability for UAV formation flight in dynamic environments. The main contributions of this paper are as follows: development of a 3D hydrodynamic flow-based collision avoidance model using a 3D doublet or ellipsoidal manifold to represent emergent dynamic obstacles, integration of the Virtual Rigid Body (VRB) formation framework to establish and maintain geometry and stability of the formation throughout the mission, and comprehensive 3D simulations validating collision-free, smooth, and computationally efficient maneuvers for both single and multi-UAV scenarios.

The rest of this paper is organized as follows. Section II proposes methods for the hydrodynamic-based collision avoidance and the VRB fomration, followed by the synthesis of those two methods and quiadcopter dynamics used in the result section. In seciton III, the hydrodynamic-based collision avoidance, VRB formations, and combinations of them are simulated to evaluate their performances under dynamic obstacle scenarios. Finally, section IV concludes the outcomes of this study and future work.

\section{Methodology}

\subsection{Hydrodynamic Flow-Based Obstacle Avoidance}

\begin{figure}[hbt!]
\centering
\includegraphics[width = 0.3\textwidth]{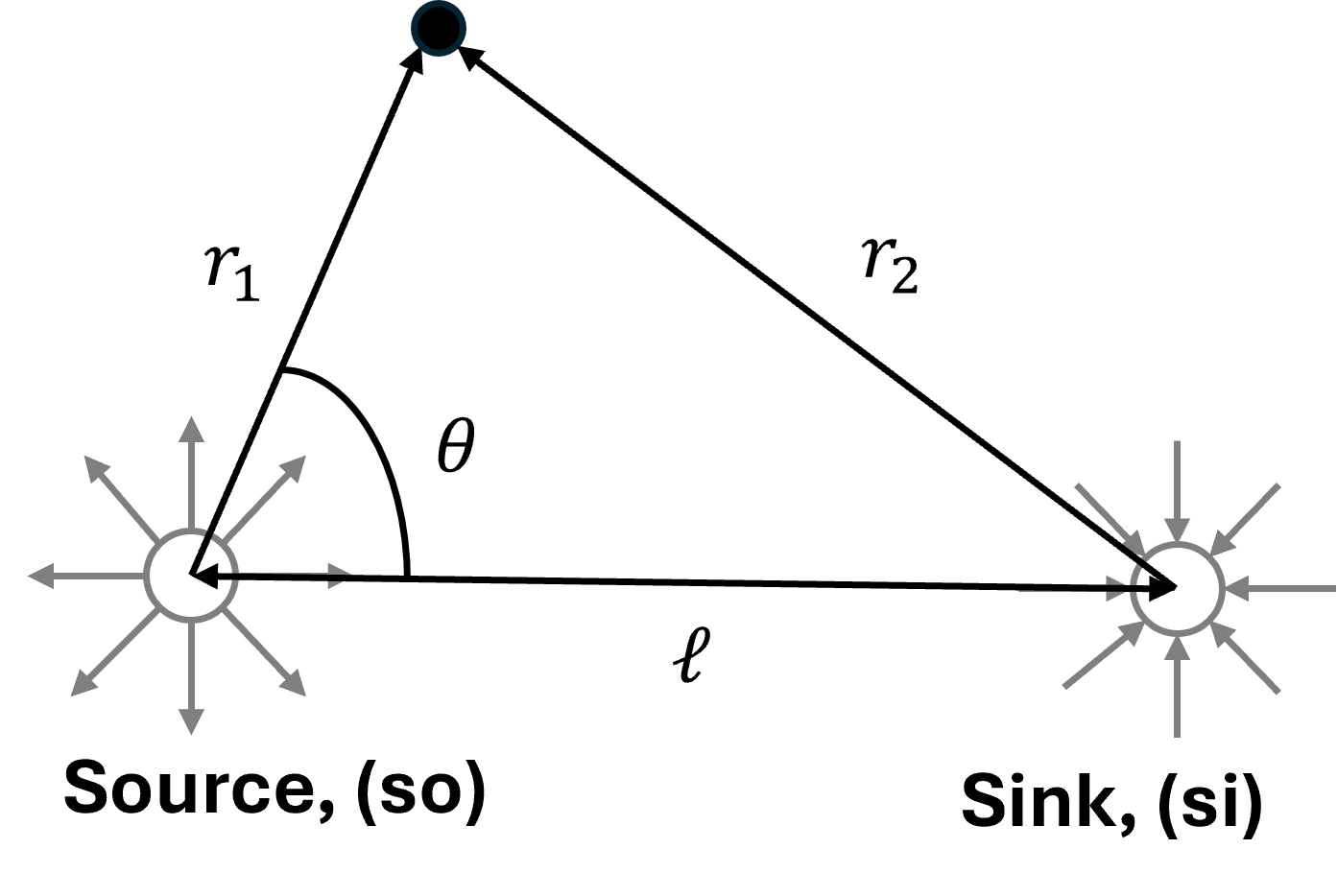}
\caption{Illustration of relationship between point source and point sink}
\label{fig:doublet}
\end{figure}

In this study, obstacles are modeled as spherical doublets or 3D Rankine oval (approximated as an ellipsoid) potential flow bodies, whose associated velocity potential satisfy Laplace's equation, thereby replicating the incompressible, irrotational, and inviscid fluid flow around solid objects for an easy yet effective flow model for real time collision avoidance. The fundamental development can be found in \cite{anderson_2016} and is extended for a moving obstacle. The expressions of a point source and point sink can be obtained from \cite{anderson_2016} as follows:
\begin{equation}
\phi_{\text{so}} = - \frac{\lambda_{\text{so}}}{4 \pi r_1}, ~~~ \phi_{\text{si}} = \frac{\lambda_{\text{si}}}{4 \pi r_2},
\label{eq:source_and_sink}
\end{equation}
where $r_1$ and $r_2$ are distances between a fluid particle and the point source and sink, respectively as illustrated in Fig.\ref{fig:doublet}, and $\lambda_{\text{so}} > 0$ and $\lambda_{\text{si}} > 0$ are the strengths of them. 

With the expressions of the point source and sink obtained, a combined flow, $\phi_{\text{c}} = \phi_{\text{so}} + \phi_{\text{si}}$, is expressed with an assumption of identical strengths (i.e. $\lambda_{\text{so}} = \lambda_{\text{si}} = \lambda_{\text{c}}$) as follows:
\begin{equation}
\phi_{\text{c}} = - \frac{\lambda_{\text{c}}}{4 \pi} \frac{r_1 - r_2}{r_1 r_2}.
\label{eq:doublet_combined_flow}
\end{equation}

Since a doublet is constructed by letting $\ell \rightarrow 0$, the low of cosines obtained from the relation in Fig.\ref{fig:doublet}, which is $r_2^2 = r_1^2 + \ell^2 - 2 r_1 \ell \cos \theta$, can be rewritten as follows:
\begin{equation}
\begin{aligned}
( r_2 + r_1 )(r_2 - r_1) = -2 r_1 \ell \cos \theta ~~~ &\because ~ \ell \rightarrow 0 \\
r_1 - r_2 = \ell \cos \theta ~~~ &\because ~ \text{as} ~ \ell \rightarrow 0, ~ r_1 \rightarrow r_2.
\end{aligned}
\end{equation}

Therefore, by letting the distance from the resulting doublet center to the fluid particle be $r$, expression of a three-dimensional doublet is obtained as follows:
\begin{equation}
\phi_{\text{c}} \rightarrow \phi_{\text{d}} = - \frac{\mu \cos \theta}{4 \pi r^2},
\label{eq:3d_doublet}
\end{equation}
where $\mu = \lambda_{\text{c}} \ell$ represents strength of the doublet. 

Since we wish to replicate the behavior of fluid flow, the velocity potential is given by Laplace's equation under the assumption of incompressible flow,
\begin{equation}
\nabla^2\phi = \frac{1}{r^2\sin\theta} \left\{ \frac{\partial}{\partial r} \left( r^2 \sin\theta \frac{\partial \phi}{\partial r} \right) + \frac{\partial}{\partial\theta} \left(\sin\theta \frac{\partial \phi}{\partial \theta}\right) + \frac{\partial}{\partial \Phi}\left(\frac{1}{\sin\theta} \frac{\partial \phi}{\partial \Phi} \right)\right\} = 0,
\end{equation}
which Eq.(\ref{eq:3d_doublet}) satisfies. 

By further assuming that the flow is irrotational, the velocity field, $\nabla \phi$, at a distance $r$ from the doublet center is found as follows:
\begin{equation}
{\bm v}_{\text{d}} = \nabla \phi_{\text{d}} = \frac{\mu \cos \theta}{2 \pi r^3} \hat{\bm e}_r + \frac{\mu \sin \theta}{4 \pi r^2} \hat{\bm e}_\theta + 0 \hat{\bm e}_\Phi.
\label{eq:doublet_velocity_field}
\end{equation}

\begin{figure}[hbt!]
	\centering
	\begin{subfigure}{0.3\textwidth}
		\centering
		\includegraphics[width = 0.8\textwidth]{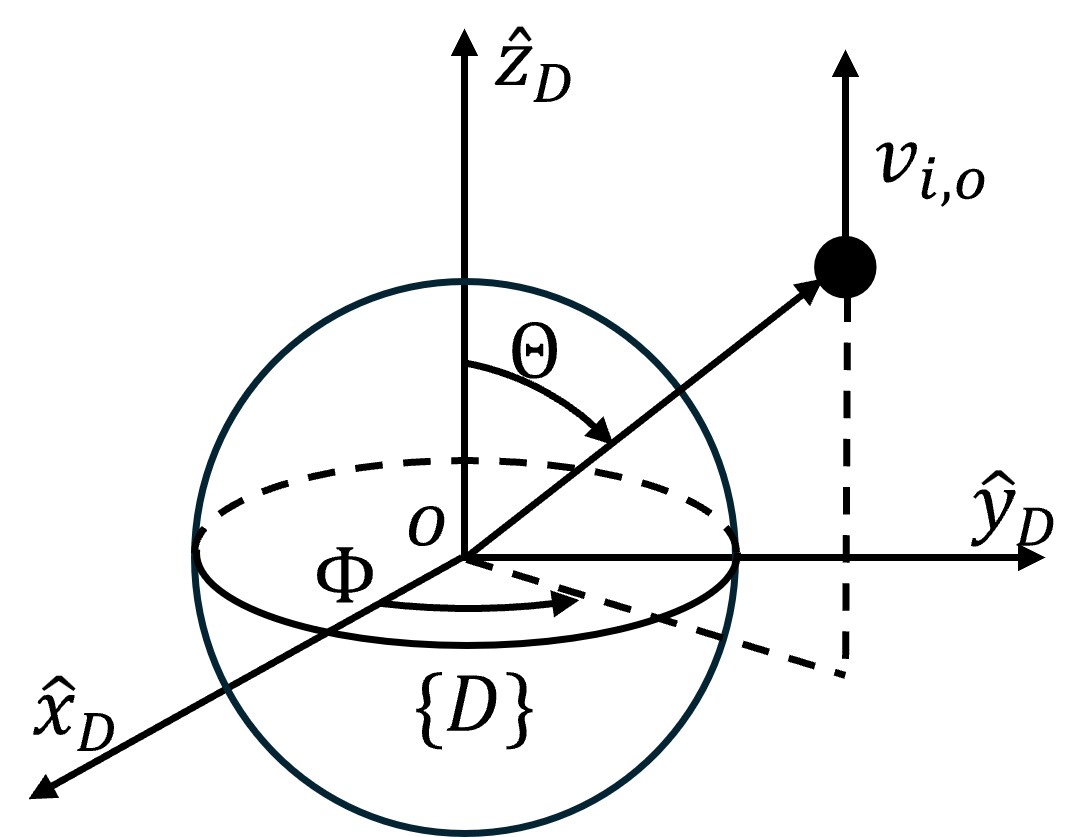}
		\caption{Illustration of the relation between the doublet manifold and a fluid particle in the doublet's local cartesian coordinate system, $\{D\}$}
		\label{fig:doublet_local}
	\end{subfigure}
	\hspace{15mm}
	\begin{subfigure}{0.3\textwidth}
		\includegraphics[width = 0.8\textwidth]{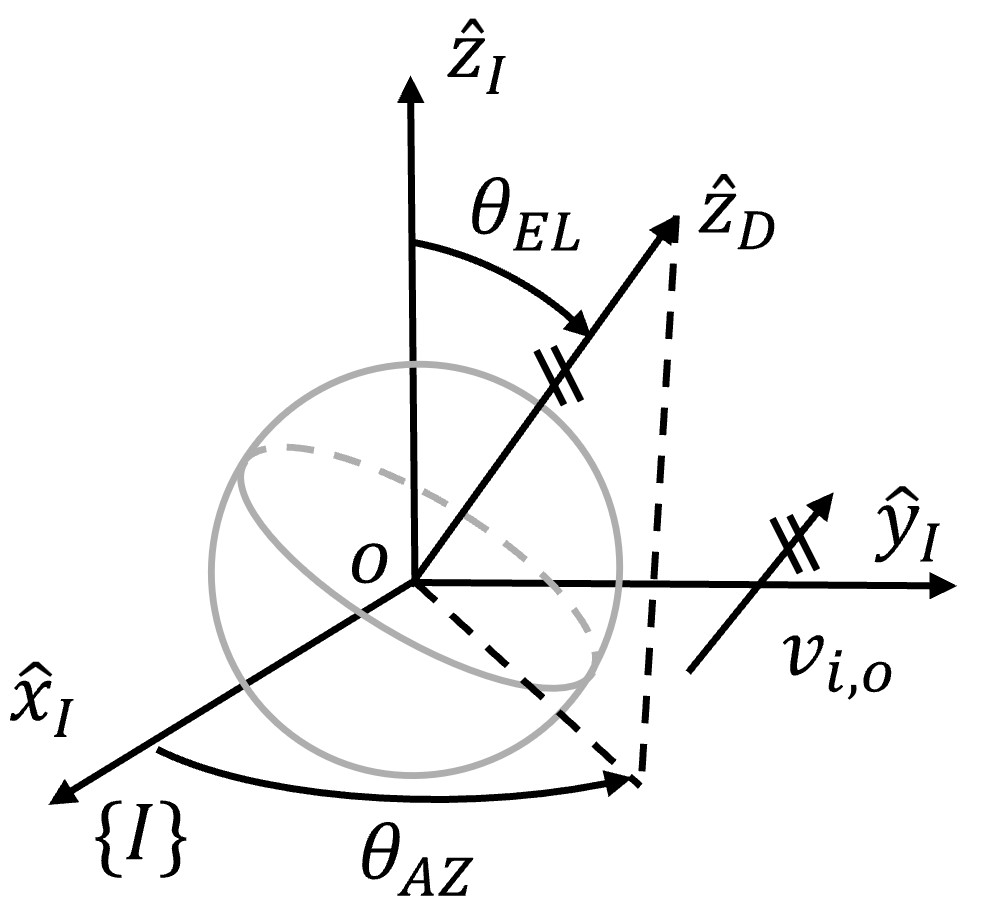}
		\caption{Illustration of the relation between the local frame $\{D\}$ and the inertial frame $\{I\}$}
		\label{fig:world_doublet}
	\end{subfigure}
	\caption{Relationship between the inertial coordinate, doublet, and fluid particle}
	\label{fig:doublet_relations}
\end{figure}

Now, let us model a flow over the doublet. The illustration of the relationship between the doublet and a fluid particle is shown in Fig.\ref{fig:doublet_relations}.

By letting a uniform velocity field of $i$-th fluid particle be ${\bm v}_i \in \mathbb{R}^{3 \times 1}( = {\bm v}_\infty)$ and the velocity of the obstacle be ${\bm v}_o \in \mathbb{R}^{3 \times 1}$, a combined flow that is expressed in doublet's spherical frame as follows:
\begin{equation}
\left\{{\bm v}_{\text{c}}\right\}_E = 
\left( v_r + \frac{\mu \cos \theta}{2 \pi r^3}  \right) \hat{\bm e}_r +
\left( v_\theta + \frac{\mu \sin \theta}{4 \pi r^3}  \right) \hat{\bm e}_\theta + 
0 \hat{\bm e}_\Phi,
\label{eq:combined_flow_local}
\end{equation}
where $v_r = \|{\bm v}_{\infty,o}\| \cos \theta$ and $v_\theta = -\|{\bm v}_{\infty,o}\| \sin \theta$ are the freestream in the spherical coordinate, and ${\bm v_{\infty,o}} = {\bm v}_\infty - {\bm v}_o~(={\bm v}_i - {\bm v}_o)$ is the relative velocity between the fluid particle and the obstacle. 

Since a flow over a body must satisfy the flow-tangency boundary condition on the body surface \cite{anderson_2016}, that is $\left\{{\bm v}_{\text{c}}\right\}_E \cdot {\bm n} = 0$, where ${\bm n}$ is a vector normal to the body's surface, the following condition is obtained:
\begin{equation}
\left\{{\bm v}_c\right\}_E \cdot {\bm n} = \left\{{\bm v}_{\text{c}}\right\}_E \cdot \hat{\bm e}_r = \| {\bm v}_{\infty,o} \| \cos \theta + \frac{\mu \cos \theta}{2 \pi r^3} = 0,
\label{eq:boundary_condition}
\end{equation}
which gives the doublet strength $\mu$ of
\begin{equation}
\mu = - 2 \pi R_d^3 \| {\bm v}_{\infty,o} \|,
\label{eq:streangth}
\end{equation}
where $R_d$ is the desired doublet radius, at which is the boundary condition, Eq.(\ref{eq:boundary_condition}), is satisfied. 

To utilize the induced flow velocity expressed in Eq.~\ref{eq:combined_flow_local} as a velocity command to a vehicle, it needs to be converted into inertial absolute velocity. The elevation angle and azimuth angle of the doublet's local cartesian coordinates with respect to the inertial frame, expressed as $\theta_{\text{EL}}$ and $\theta_{\text{AZ}}$ in Fig.~\ref{fig:doublet_relations}, are calculated as follows:
\begin{equation}
\theta_{\text{EL}} = \cos^{-1}\left\{ \frac{\{{\bm v}_{\infty,o}\}_{D,z}}{\| {\bm v}_{\infty,o} \|} \right\},
\end{equation}
\begin{equation}
\theta_{\text{AZ}} = \tan^{-1} \left\{ \frac{\{{\bm v}_{\infty,o}\}_{D,y}}{\{{\bm v}_{\infty,o}\}_{D,x} } \right\}.
\end{equation}

Then, the transformation matrix from the inertial frame, $\{I\}$, to the doublet's local cartesian coordinates, $\{D\}$, is calculated as follows:
\begin{equation}
{\bm T}_I^D = 
\begin{bmatrix}
\cos \theta_{EL} & 0 & -\sin \theta_{EL} \\
0 & 1 & 0 \\
\sin \theta_{EL} & 0 & \cos \theta_{EL}
\end{bmatrix}
\begin{bmatrix}
\cos \theta_{AZ} & \sin \theta_{AZ} & 0 \\
-\sin \theta_{AZ} & \cos \theta_{AZ} & 0 \\
0 & 0 & 1
\end{bmatrix}
=
\begin{bmatrix}
\cos \theta_{EL} \cos \theta_{AZ} & \cos \theta_{EL} \sin \theta_{AZ} & - \sin \theta_{EL} \\
- \sin \theta_{AZ} & \cos \theta_{AZ} & 0 \\
\sin \theta_{EL} \cos \theta_{AZ} & \sin \theta_{EL} \sin \theta_{AZ} & \cos \theta_{EL}
\end{bmatrix}.
\label{eq:ItoD}
\end{equation}

Given the induced combined flow velocity in Eq.~(\ref{eq:combined_flow_local}) is in the doublet's spherical coordinate system, $\{E\}$, $\theta$ and $\Phi$ needs to be obtained to express it in the doublet's local cartesian coordinate system, $\{D\}$. Firstly, the instantaneous relative position between the vehicle and the obstacle, ${\bm r}_{i,o}$, can be converted into the doublet local frame as follows:
\begin{equation}
\left\{ {\bm r}_{i,o} \right\}_D = T_I^D \left\{ {\bm r}_{i,o} \right\}_I.
\end{equation}

Then, the angles $\theta$ and $\Phi$ depicted in Fig.~\ref{fig:doublet_local} can be found as follows.:
\begin{equation}
\theta = \cos^{-1}\left\{ \frac{\{{\bm r}_{i,o}\}_{D,z}}{\| {\bm r}_{i,o} \|} \right\}, 
\end{equation}
\begin{equation}
\Phi = \tan^{-1} \left\{ \frac{\{{\bm r}_{i,o}\}_{D,y}}{\{{\bm r}_{i,o}\}_{D,x}}  \right\}.
\end{equation}

Then, the transformation matrix from doublet's spherical coordinate system to its local cartesian coordinate system can be constructed as follows:
\begin{equation}
{\bm T}_E^D = 
\begin{bmatrix}
\sin \theta \cos \Phi & \cos \theta \cos \Phi & - \sin \Phi \\
\sin \theta \sin \Phi & \cos \theta \sin \Phi & \cos \Phi \\
\cos \theta & - \sin \theta & 0
\end{bmatrix}.
\label{eq:EtoD}
\end{equation}

Finally, the combined velocity field in the doublet's local spherical coordinate given in Eq.~(\ref{eq:combined_flow_local}) can be transformed into the absolute velocity as follows:
\begin{equation}
{\bm v}_{i,h} = \left\{{\bm v}_o\right\}_I + \left[ T_I^D \right]^T \left[ T_E^D \right] \left\{ {\bm v}_c \right\}_E
\end{equation}

When there exist multiple obstacles within an environment, the same procedures given after Eq.~(\ref{eq:combined_flow_local}) onward are repeated for other obstacles by regarding ${\bm v}_{i,h}$ as the uniform flow, ${\bm v}_\infty$, thanks to the superposition characteristics of the harmonic function. 

Once the induced flow velocity field generated by all obstacles within the sensing region is computed, it is used as the reference velocity for the vehicle. While various velocity-tracking strategies may be adopted, we will employ a simple proportional controller,
\begin{equation}
{\bm f}_{i,h} = K_h ( {\bm v}_{i,h} - {\bm v}_i ),
\end{equation}
to directly evaluate the feasibility and fundamental behavior of the hydrodynamic-flow based avoidance mechanism. This formulation enables the vehicle to align its motion with the local flow field and execute smooth and continuous maneuvers without explicit trajectory replanning.

\subsection{Virtual Rigid Body Formation}
While the hydrodynamic-flow based model provides a smooth and reactive mechanism for avoiding emergent obstacles, a strategy is additionally required to ensure unified and cohesive motion among multiple UAVs. Therefore, the Virtual Rigid Body (VRB) formation framework, one of the virtual structure formation approaches, is introduced as the method for achieving multi-UAV coordinated operation due to its ease in formation maintenance and guidance. 

\begin{figure}[htb!]
\centering
\includegraphics[width=0.3\textwidth]{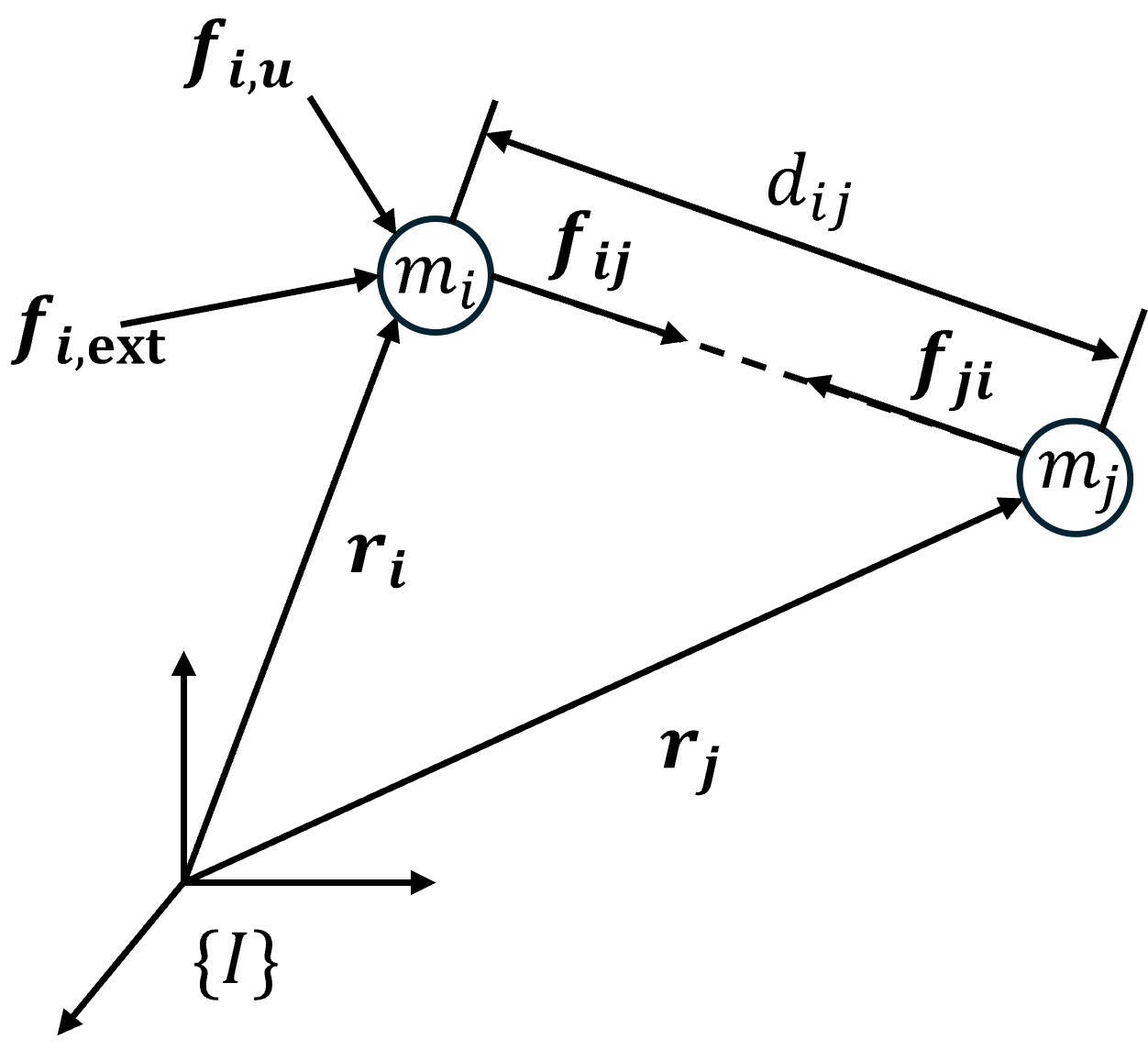}
\caption{Illustration of the relationship between agents}
\label{fig:agents_relation}
\end{figure}

The VRB formation utilizes the concept of \textit{constraint forces}, early introduced in \cite{witkin_1990}, which constrain the motion of agents in the system to maintain inter-agent distance constraints. The development of the constraint forces and the VRB formation is well explained in \cite{sato_2024, zou_2007}. The illustration of the relation between agents are shown in Fig.~\ref{fig:agents_relation}.

When a multi-agent system is described as shown in Fig.~\ref{fig:agents_relation}, the total force acting on the $i$-th agent in the system can be expressed as follows \cite{junkins_2021}:
\begin{equation}
{\bm f}_i = m_i \ddot{\bm r}_i = {\bm f}_{i,\text{ext}} + {\bm f}_{i,u} + \sum_{j=1, j \neq i}^{N-1} {\bm f}_{ij}
\label{eq:multi_agent_force}
\end{equation}
where $m_i$ is the mass of $i$-th agent, ${\bm f}_{i,\text{ext}} \in \mathbb{R}^{3 \times 1}$ is the external forces, such as gravitational force and aerodynamic forces, ${\bm f}_{i,u} \in \mathbb{R}^{3 \times 1}$ is the control input for commanded trajectory and velocity tracking, and ${\bm f}_{ij} \in \mathbb{R}^{3 \times 1}$ is the inter-agent force between $i$-th and $j$-th agents, which is later utilized as the constraint forces. 

The VRB formation is established and maintained through the constraint forces, which, in a stable manner, satisfies the constraint,
\begin{equation}
c_k = \| {\bm r}_i - {\bm r}_j \| - d_{ij,d},
\label{eq:constraint_function}
\end{equation}
where ${\bm r}_i \in \mathbb{R}^{3 \times 1}$ and ${\bm r}_j$ are position vectors of $i$-th and $j$-th agent, respectively, and $d_{ij,d}$ is the desired distance between them. 

To achieve the constraint, Eq.~(\ref{eq:constraint_function}), in a stable manner, the following constraints need to be satisfied as well:
\begin{equation}
\dot{\bm c} = {\bm J}\dot{\bm r} = 0, 
\label{eq:velocity_constraint}
\end{equation}
\begin{equation}
\ddot{\bm c} = \dot{\bm J}\dot{\bm r} + {\bm J}\ddot{\bm r} = 0,
\label{eq:acceleration_constraint}
\end{equation}
where ${\bm J} = \left[ \frac{\partial {\bm c}}{\partial {\bm r}} \right]$, ${\bm c} = \left[ c_1,~c_2, \ldots, \ldots, c_m \right]^T$, and ${\bm r} = \left[ {\bm r}_1^T, ~{\bm r}_2^T,~\ldots, ~{\bm r}_N^T \right]^T \in \mathbb{R}^{3N \times 1}$. Additionally, there exists $N(N-1)/2$ total numbers of distance constraints in a $N$-agent system. Among those constraints, the number of distance constraints, which when satisfied the formation becomes rigid, is $m = 3N-6~~(N \geq 3)$ for three-dimensional motion. For an $N$-particle system in three-dimensional space, the total number of coordinates is 3$N$. The motion of the system as a whole is characterized by 3 translational and 3 rotational degrees of freedom (DoF). Since the DoF are defined as
\begin{equation}
\text{DoF} = \text{number of coordinates} - \text{number of independent constraints},
\end{equation}
the system has $3N-6$ internal DoF. If these $3N-6$ DoF are fully constrained, the structure becomes rigid. Therefore, $3N-6$ independent constraints are required to ensure rigidity.

Now, by substituting $\ddot{\bm r}$ in Eq.~(\ref{eq:acceleration_constraint}) into that in the collective form of Eq.~(\ref{eq:multi_agent_force}); then, by solving resulting expression for $\sum_{j=1, j \neq i}^{N-1} {\bm f}_{ij} (= {\bm f}_{i,c})$, the constraint force that satisfies Eq.~(\ref{eq:acceleration_constraint}) can be obtained as follows:
\begin{equation}
{\bm f}_c = \left[ {\bm J}{\bm M}^{-1} \right]^{-1}\left[ - \dot{\bm J}\dot{\bm r} - {\bm J} {\bm M}^{-1} ({\bm f}_{\text{ext}} + {\bm f}_u) \right],
\label{eq:constraint_force_from_acceleration}
\end{equation}
where ${\bm f}_c = [{\bm f}_{1,c}^T, ~{\bm f}_{2,c}^T, ~\ldots, ~{\bm f}_{N,c}^T] \in \mathbb{R}^{3N \times 1}$, ${\bm f}_{\text{ext}(u)} = \left[ {\bm f}_{1,\text{ext}(u)}^T, ~{\bm f}_{2,\text{ext}(u)}^T, ~\ldots, ~{\bm f}_{N,\text{ext}(u)}^T \right]^T \in \mathbb{R}^{3N \times 1}$, and ${\bm M} = \text{diag}(m_1, ~m_2, ~\ldots, ~m_N) \otimes {\bm I}_{3 \times 3} \in \mathbb{R}^{3N \times 3N}$. 

To satisfy Eq.~(\ref{eq:velocity_constraint}) as well, d'Alembert's principle of virtual work, which states that, given a smooth holomonic constraint, the constraint force is always normal to the constraint surface (i.e. the constraint force is in the direction of the constraint surface gradient, ${\bm J}$) \cite{junkins_2021}, is introduced. Then, by utilizing the constraint sensitivity parameter ${\bm \lambda}$ (i.e. Lagrange multiplier), the constraint force is expressed as follows:
\begin{equation}
{\bm f}_c = {\bm J}^T {\bm \lambda}.
\label{eq:virtual_work}
\end{equation}

Additionally, the principle states that the virtual work done by the constraint force associated with the holonomic constraints is zero (i.e. ${\bm f}_{i,c} \delta{\bm r}_i = 0$) \cite{junkins_2021}. This can be extended to the so-called virtual power (i.e. ${\bm f}_{i,c}\delta\dot{\bm r}_i = 0$) \cite{Moon_1998}, which corresponds to the condition of interest in Eq.~(\ref{eq:velocity_constraint}). Given this characteristic, by substituting ${\bm f}_c$ in Eq.~(\ref{eq:constraint_force_from_acceleration}) into that in Eq.~(\ref{eq:virtual_work}), the sensitivity parameter, ${\bm \lambda}$, can be obtained. Further, by substituting the resulting ${\bm \lambda}$ back into Eq.~(\ref{eq:virtual_work}), expression of the constraint force that satisfies both conditions given in Eq.~(\ref{eq:velocity_constraint}) and Eq.~(\ref{eq:acceleration_constraint}) is obtained as follows:
\begin{equation}
{\bm f}_c = {\bm J}^T \left[ {\bm J} {\bm M}^{-1} {\bm J}^T \right]^{-1} \left[ - \dot{\bm J}\dot{\bm r} - {\bm J}{\bm M}^{-1} ({\bm f}_{\text{ext}} + {\bm f}_u) \right]
\label{eq:constraint_force_comp}
\end{equation}

Since no control law is embedded in Eq.~(\ref{eq:constraint_force_comp}), initial conditions need to be carefully chosen to satisfy the constraint functions \cite{ihle_2005}. This becomes increasingly impractical and tedious as the number of agents increases. Thus, we will employ a PID-like feedback control law inherited from the Baumgarte stabilization technique \cite{baumgarte_1972}, which is often utilized for numerical stabilization of multi-agent constrained systems. As a result, the following finalized form of the constraint force is developed \cite{sato_2024}:
\begin{equation}
{\bm f}_c = {\bm J}^T \left[ {\bm J}{\bm M}^{-1} {\bm J}^T \right]^{-1}
\left[
-{\bm J}{\bm M}^{-1}\left({\bm f}_{\text{ext}} + {\bm f}_u \right) - \dot{\bm J}\dot{\bm r} 
\underbrace{-2\alpha\dot{\bm c} - \beta^2{\bm c} - \gamma \int {\bm c} dt}_\text{stabilization term}
\right],
\label{eq:fc_final}
\end{equation}
where $\alpha$, $\beta$, and $\gamma$ are tuning parameters for the feedback controller, and $\alpha = \beta$.

\begin{figure}[htb!]
\centering
\includegraphics[width = 0.3\textwidth]{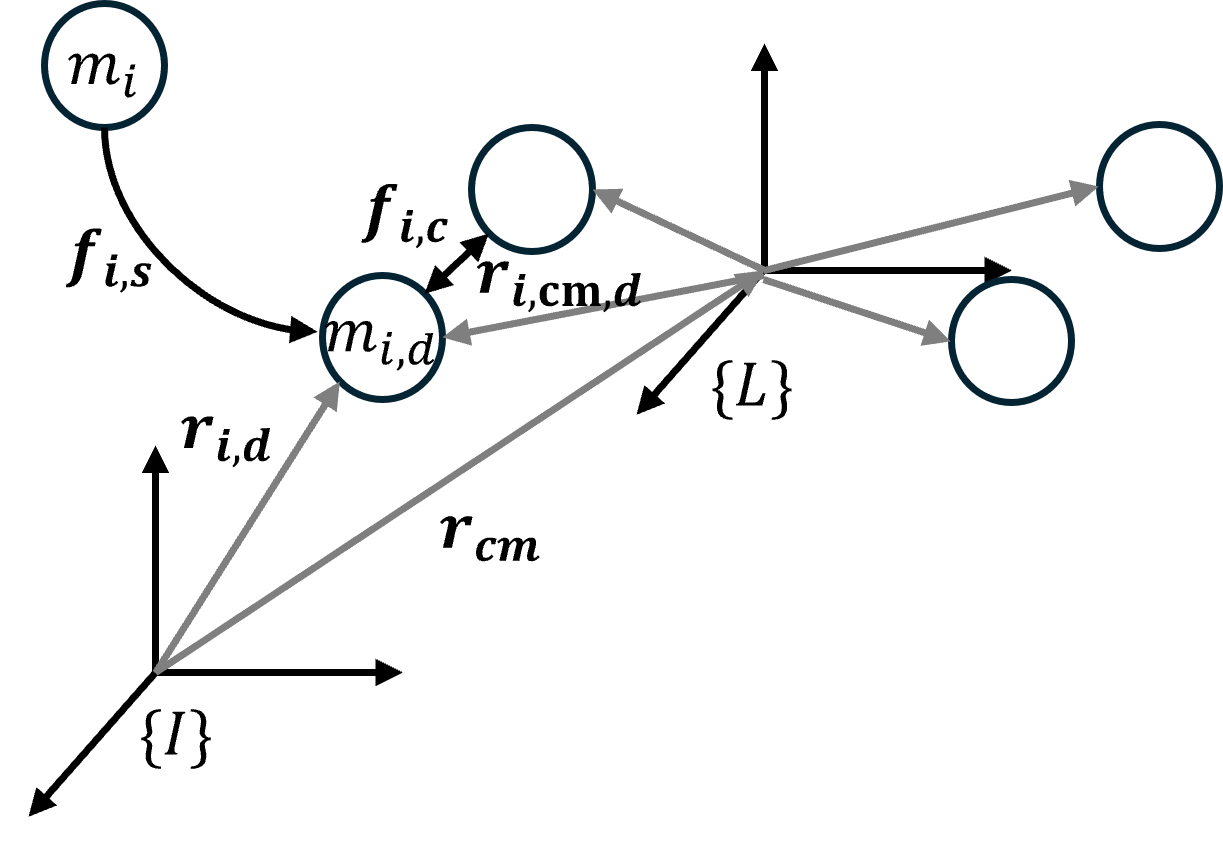}
\caption{Illustration of the agent slot allocation method}
\label{fig:slot_allocation}
\end{figure}

While satisfying the $3N-6$ distance constraints is a sufficient condition to ensure structural rigidity, it does not guarantee that the rigid structure achieves the desired geometry and orientation. Consequently, the slot allocation procedure is necessary to ensure that the realized rigid body structure not only remains rigid but also conforms to the prescribed geometry and orientation. The slot for each agent is selected based on the cumulative distance for them to travel to fill in those slots. Illustration of this scheme is provided in Fig.~\ref{fig:slot_allocation}.

The force ${\bm f}_{i,s}$ represents control input applied to $i$-th agent to reach the reference position,
\begin{equation}
{\bm r}_{i,d} = {\bm r}_{\text{cm}} + {\bm T}_L^I {\bm r}_{i,\text{cm},d}.
\end{equation}

A remaining challenge is how to map the local slot positions, ${\bm r}_{i,\text{cm},d}$ to agents. Different allocation may require different travel distances for the agents to reach their designated positions, which would result in affecting transient control effort and formation establishment time. To tackle this challenge, an optimal slot allocation procedure is formulated. By letting ${\bm S} \in \mathbb{R}^{3 \times N}$ be a matrix where each column ${\bm S}(:,\text{col})$ stores local slot positions of the desired formation, and ${\bm \Pi}_N$ be the set of all $N!$ sequence of $N$ slots. We seek an optimal sequence of slot positions $\pi^* \in {\bm \Pi}_N$ that minimizes the total distance between each agent and its assigned slot:
\begin{equation}
\pi^* = \arg \min_{\pi \in \Pi_N} \sum_{i=1}^N \| {\bm r}_{i,\text{cm}} - {\bm S}(:, \pi(i)) \|_2.
\label{eq:total_min_slot_allocation}
\end{equation}

Once the optimal slot sequence $\pi^*$ is determined, the slot allocation can be done as follows:
\begin{equation}
{\bm r}_{i,\text{cm},d} = {\bm S}(:,\pi^*(i)).
\end{equation}

\subsection{Algorithm Synthesis}
To integrate the hydrodynamic-flow based obstacle avoidance mechanism and the VRB formation framework into a unified multi-UAV architecture, the individual components must be synthesized into a cohesive algorithm that governs trajectory tracking, obstacle avoidance, and formation establishment and maintenance. The fundamental limitation of traditional potential field approaches is the absence of a defined trajectory for the vehicles to follow. To address this, the overall commanded control input for the $i$-th vehicle is defined as
\begin{equation}
{\bm f}_{i,\text{cmd}} = {\bm f}_{i,h} + {\bm f}_{i,s} + {\bm f}_{i,g} + {\bm f}_{i,c},
\end{equation}
where ${\bm f}_{i,h}$ is the hydrodynamic-flow induced velocity tracking force, ${\bm f}_{i,s}$ is the slot allocation control input for formation geometry, ${\bm f}_{i,g}$ compensates the gravitational force, and ${\bm f}_{i,c}$ the constraint force that maintains desired inter-agent distances.

In a practical point of view, it is reasonable to consider ${\bm f}_{i,h}$ only when there are obstacles within a vehicle's sensing range. Taking this into account, the overall algorithm of this entire scheme shown in Algorithm \ref{alg:algorithm} is proposed. The combined result yields the total control input ${\bm f}_{i,\text{cmd}}$ for each vehicle, which is subsequently mapped into thrust and attitude commands through the quadcopter dynamics described in Subsection D.

This unified synthesis enables smooth and coordinated multi-UAV motion in dynamic environments, preserving formation structure while reacting to emergent obstacles in real time.

\subsection{Quadcopter Dynamics}
\begin{figure}[htb!]
\centering
\includegraphics[width = 0.3\textwidth]{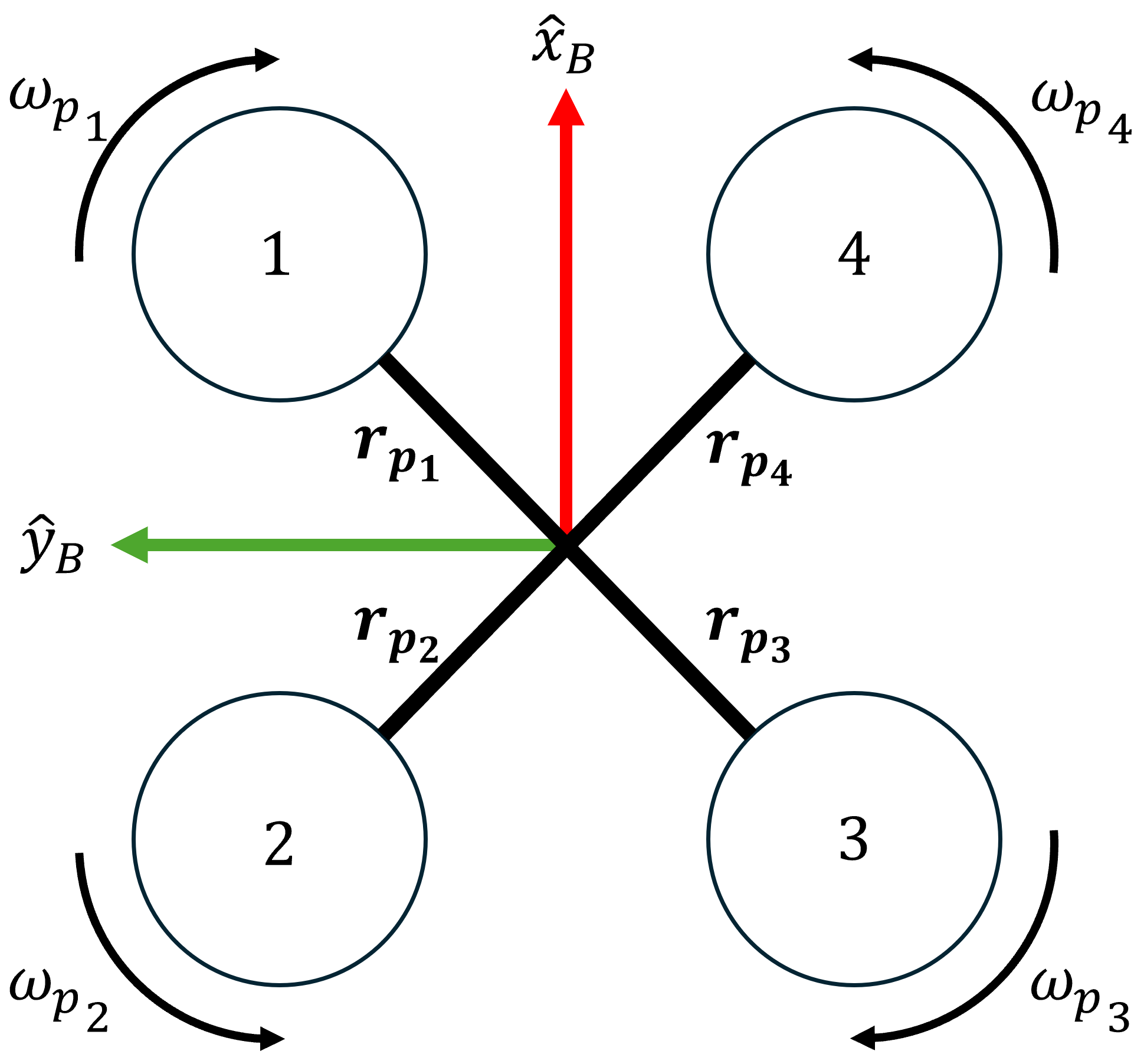}
\caption{Quadcopter diagram}
\label{fig:quadcopter_diagram}
\end{figure}

In the subsequent sections, we will consider quadcopters as the swarm agents. The dynamics of a quadcopter is expressed as follows:
\begin{equation}
\dot{\bm v}_i = - {\bm \omega}_i \times {\bm v}_i + \frac{T_i}{m_i}{\bm e}_3 - {\bm T}_I^B g {\bm e}_3
\end{equation}
\begin{equation}
\dot{\bm r}_i = {\bm T}_B^I {\bm v}_i
\end{equation}
\begin{equation}
\dot{\bm \omega}_i = - {\bm I}_{i}^{-1} \Big( {\bm \omega}_i \times {\bm I}_i {\bm \omega}_i - {\bm \tau}_i \Big)
\end{equation}
\begin{equation}
\dot{\bm q}_i = \frac{1}{2}{\bm B}({\bm \omega}_i) {\bm q}_i
\end{equation}
where ${\bm v}_i \in \mathbb{R}^{3 \times 1}$ is the velocity vector of the $i$-th quadcopter expressed in its body-fixed frame, ${\bm \omega}_i$ is the vehicle angular rate, $T_i$ is the total thrust magnitude, ${\bm T}_I^B$ is the inertial-to-body transformation matrix, ${\bm e}_3 = [0, ~0, ~1]^T$, ${\bm I}_i$ is the moment of the vehicle, ${\bm \tau}_i$ is the torque acting on the vehicle, and ${\bm q}_i$ is the quadcopter's scalar-first quaternions. 

The inertial-to-body frame transformation matrix can be obtained by the following:
\begin{equation}
{\bm T}_I^B = 
\begin{bmatrix}
q_0^2 + q_1^2 - q_2^2 - q_3^2 & 2 ( q_1 q_2 + q_0 q_3 ) & 2 ( q_1 q_3 - q_0 q_2 ) \\
2 ( q_1 q_2 - q_0 q_3 ) & q_0^2 - q_1^2 + q_2^2 - q_3^2 & 2 ( q_2 q_3 + q_0 q_1 ) \\
2 ( q_1 q_4 + q_0 q_2 ) & 2 ( q_2 q_3 - q_0 q_1 ) & q_0^2 - q_1^2 - q_2^2 + q_3^2
\end{bmatrix}_i
\end{equation}

Meanwhile applying a control input to an unconstrained mass particle is easily done, it needs to be converted into an appropriate thrust and attitude command \cite{Zuo_2010} when a quadcopter is considered as an agent. By letting ${\bm f}_{i,\text{cmd}} = {\bm u}_i = [u_{i,x}, ~u_{i,y}, ~u_{i,z}]^T$ be the inertial control input that is required for $i$-th vehicle. Then, the thrust, pitch, and roll commands are found as follows:
\begin{equation}
\begin{aligned}
T_i = m_i 
&~[~u_{i,x} ( \sin \theta_i \cos \psi_i \cos \phi_i + \sin \psi_i \sin \phi_i ) \\
&+ u_{i,y} ( \sin \theta_i \sin \psi_i \cos \phi_i - \cos \psi_i \sin \phi_i ) \\
&+ u_{i,z} \cos \theta_i \cos \phi_i ~]
\end{aligned}
\end{equation}
\begin{equation}
\theta_{i,\text{cmd}} = \tan^{-1} \left( \frac{ u_{i,x} \cos \psi_{i,\text{cmd}} + u_{i,y} \sin \psi_{i,\text{cmd}} }{u_{i,z}} \right)
\end{equation}
\begin{equation}
\phi_{i,\text{cmd}} = \sin^{-1} \left( \frac{ u_{i,x} \sin \psi_{i,\text{cmd}} - u_{i,y} \cos \psi_{i,\text{cmd}} }{ \| {\bm u}_i \|_2 } \right)
\end{equation}
where $\phi_i$, $\theta_i$, and $\psi_i$ are roll, pitch, and yaw of the $i$-th vehicle, respectively, and $\phi_{i,\text{cmd}}$, $\theta_{i,\text{cmd}}$, and $\psi_{i,\text{cmd}}$ are commanded roll, pitch, and yaw, respectively. The PID controller is employed for both outer- and inner-loop of all vehicles. 

The vehicle Euler angles can be obtained from the entities of ${\bm T}_I^B$ as follows:
\begin{equation}
\phi_i = \tan^{-1} \left( \frac{{\bm T}_I^B[3,3]}{{\bm T}_I^B[2,3]} \right)
\end{equation}
\begin{equation}
\theta_i = - \sin^{-1} \left( {\bm T}_I^B[1,3] \right)
\end{equation}
\begin{equation}
\psi_i = \tan^{-1} \left( \frac{{\bm T_I^B[1,1]}}{{\bm T}_I^B[1,2]} \right)
\end{equation}
where ${\bm T}_I^B[m,n]$ is $m$-th row $n$-th column entry of ${\bm T_I^B}$ matrix.

With the thrust, $T_i$, and commanded torque, ${\bm \tau}_{i,\text{cmd}}$, the calculation shown in Eq.(\ref{eq:prop_omega}) is carried out to obtain angular velocities of the propellers of a quadcopter. A diagram of a quadcopter is provided in Fig.\ref{fig:quadcopter_diagram}.

\begin{equation}
{\bm \omega}_{p,i}^2 = {\bm A}^{-1} 
\begin{bmatrix}
T_i \\
{\bm \tau}_{i,\text{cmd}}
\end{bmatrix},
\label{eq:prop_omega}
\end{equation}
where 
\begin{equation}
{\bm A} = 
\begin{bmatrix}
k_{t,1} & k_{t,2} & k_{t,3} & k_{t,4} \\
({\bm r}_{p,1})_y k_{t,1} & ({\bm r}_{p,2})_y k_{t,2} & ({\bm r}_{p,3})_y k_{t,3} & ({\bm r}_{p,4})_y k_{t,4} \\
-({\bm r}_{p,1})_x k_{t,1} & -({\bm r}_{p,2})_x k_{t,2} & -({\bm r}_{p,3})_x k_{t,3} & -({\bm r}_{p,4})_x k_{t,4} \\
k_{d,1} & -k_{d,2} & k_{d,3} & -k_{d,4}
\end{bmatrix},
\label{eq:prop_allocation_matrix}
\end{equation}
${\bm r}_{p,i}$ is the position of aerodynamic center of $i$-th propeller relative to the quadcopter's center of gravity, and $k_{t,i}$ and $k_{d,i}$ are the thrust and moment coefficient of the propeller, respectively.

\section{Simulation Results}

\begin{figure}[hbt!]
	\centering
	\begin{subfigure}{0.3\textwidth}
		\centering
		\includegraphics[width = 1.0\textwidth]{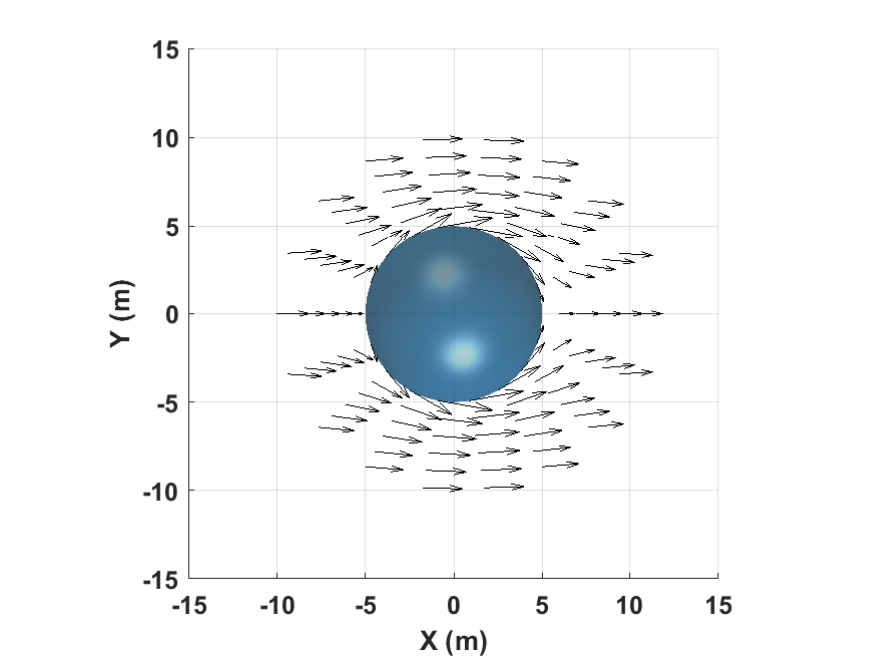}
		\caption{Velocity field around a 3D doublet}
		\label{fig:vel_field}
	\end{subfigure}
	\begin{subfigure}{0.3\textwidth}
		\includegraphics[width = 1.0\textwidth]{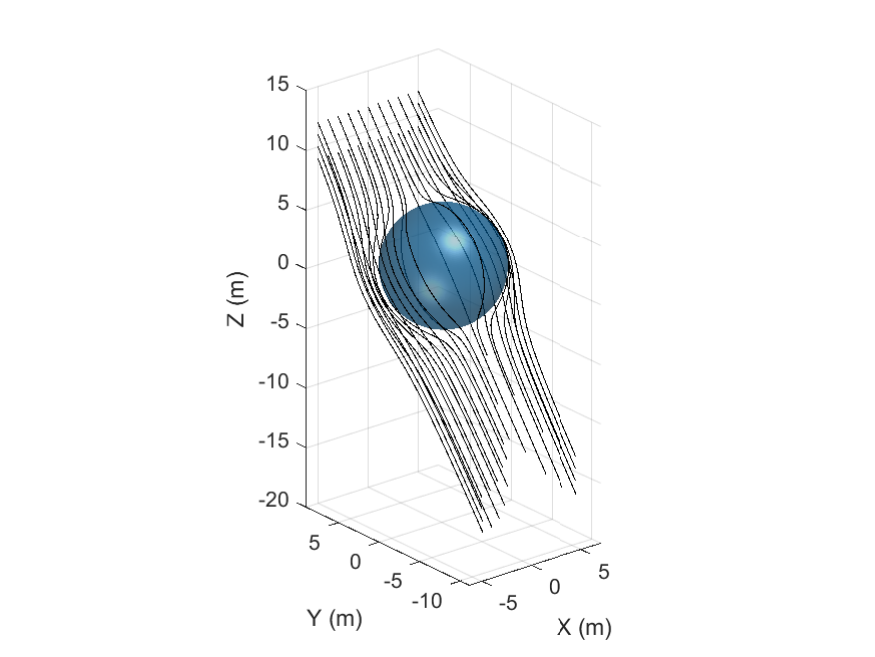}
		\caption{Fluid flow over a 3D doublet}
		\label{fig:streamline3D}
	\end{subfigure}
	\begin{subfigure}{0.39\textwidth}
		\includegraphics[width = 1.0\textwidth]{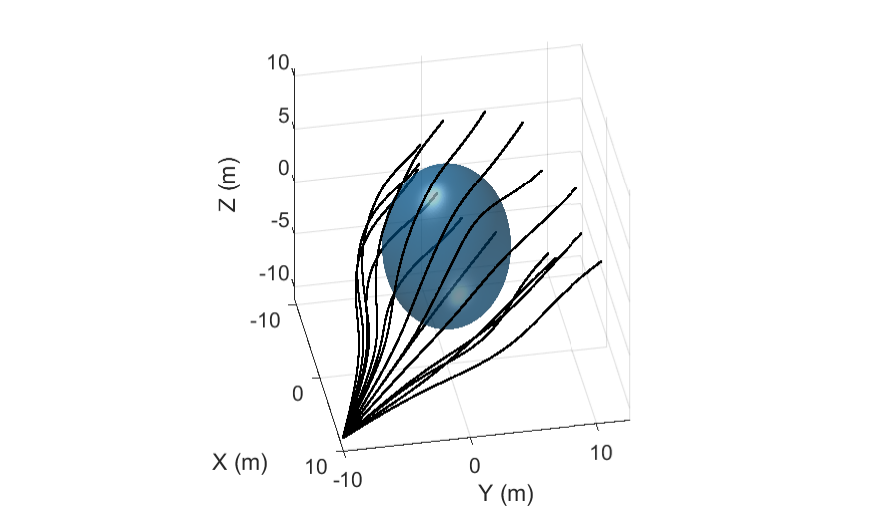}
		\caption{Fluid flow over an ellipsoid manifold}
		\label{fig:streamline_ellipsoid}
	\end{subfigure}
	\caption{Fluid flow simulation over a 3D doublet and ellipsoidal manifold}
	\label{fig:streamline}
\end{figure}

To evaluate performance and feasibility of the proposed methods, several simulations are conducted based on the strategies described in the previous section. These simulations demonstrate how UAVs, either in the individual operation or in a coordinated formation, can avoid moving obstacles using the 3D doublet flow model while maintaining trajectory tracking and formation. The simulation environment assumes ideal sensing to isolate the behavior of the core algorithm. Supplemental videos of the subsequent simulations can be found in the following link: \url{https://youtu.be/QcZWW4lSVJo}. To begin with, a velocity field as well as fluid flows around a 3D doublet are visualized as shown in Fig.~\ref{fig:streamline} to provide a visual interpretation of the model. Moreover, by repeating the same process provided in Subsection A of Section II for a source-sink flow instead of a doublet, three-dimensional flow over an ellipsoid can be also developed with proper boundary conditions as shown in Fig.~\ref{fig:streamline_ellipsoid}.

\begin{figure}[hbt!]
	\centering
	\begin{subfigure}{0.45\textwidth}
		\centering
		\includegraphics[width = 0.7\textwidth]{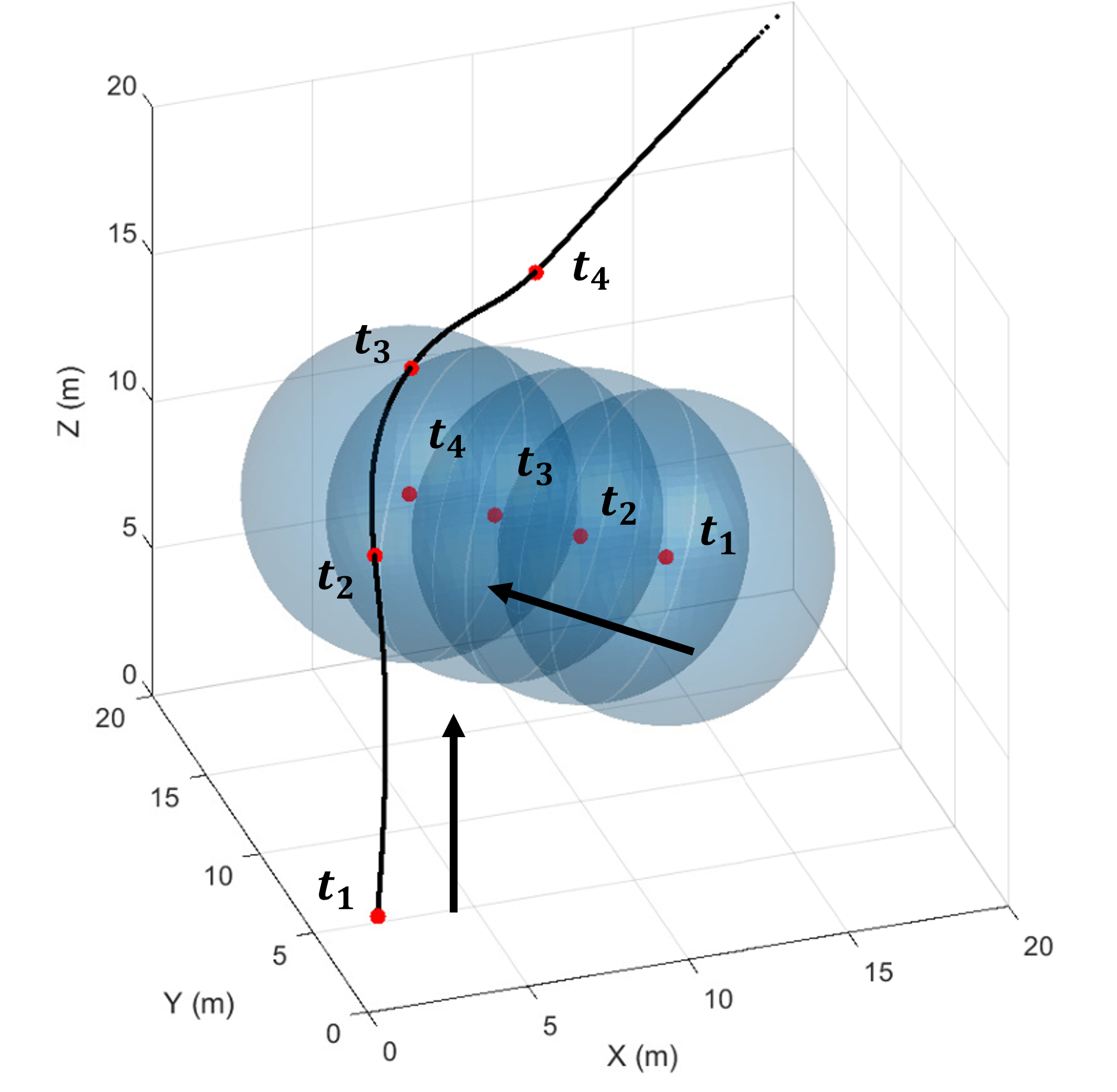}
		\caption{Point mass trajectory over a moving doublet}
		\label{fig:point_mass_doublet}
	\end{subfigure}
	\begin{subfigure}{0.45\textwidth}
		\includegraphics[width = 0.8\textwidth]{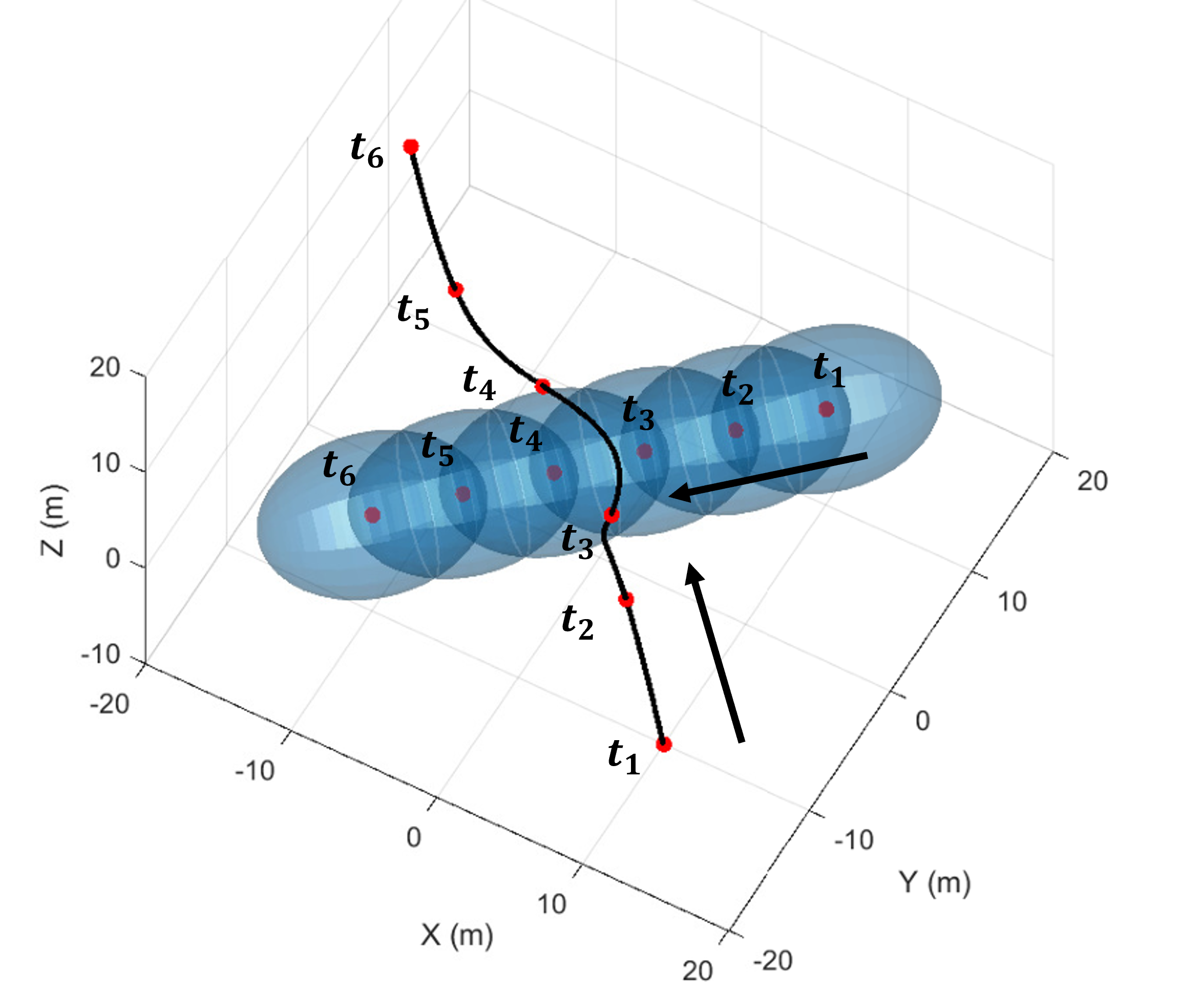}
		\caption{Point mass trajectory over a moving ellipsoid}
		\label{fig:point_mass_ellipsoid}
	\end{subfigure}
	\caption{Point mass trajectory over a 3D doublet and ellipsoidal manifold}
	\label{fig:point_mass_trajectoies}
\end{figure}

Furthermore, trajectories of unconstrained point mass avoiding a moving doublet and ellipsoidal manifold are simulated and visualized as shown in Fig.~\ref{fig:point_mass_trajectoies}. 

As Fig.~\ref{fig:point_mass_trajectoies} and the supplemental videos show, the point mass avoided the moving doublet and ellipsoidal manifolds with smooth manner successfully. Given these results, obstacle avoidance by single quadcopter is carried out in the next section. In the subsequent sections, the vehicle and simulation parameters \cite{liu_2022} given in Table~\ref{tab:vehicle_spec} are considered for the quiadcopter model.

\begin{table}[hbt!]
\caption{Quadcopter and Simulation Specifications}
\centering
\begin{tabular}{lcccccc}
\hline
\textbf{Parameter} & \textbf{Value}
\\\hline
Gravitational acceleration, g& 9.81 \(m/s^2\)\\
Mass, m				    	 & 1.023 kg    	 \\
Arm length, b				 & 0.2223 m      \\
Thrust coefficient, \(k_T\)	 & 1.4865\(\times\)10\(^-7\) N/(rad/s)\(^2\) \\
Moment coefficient, \(k_d\)  & 2.9250\(\times\)10\(^{-9}\) N\(\cdot\)m/(rad/s)\(^2\) \\
Sensing range	& 10 m in radius \(\times\) \(\pm\) 3 m in height\\
Inertia properties:			 & 						  \\
\(I_{xx}\)					 & 0.0095 \(kg\cdot m^2\) \\
\(I_{yy}\)					 & 0.0095 \(kg\cdot m^2\) \\
\(I_{zz}\)					 & 0.0186 \(kg\cdot m^2\) \\
\hline
\label{tab:vehicle_spec}
\end{tabular}
\end{table} 

Additionally, to evaluate robustness of the proposed method against noise, Gaussian noise is added to the positions of obstacles within the vehicle's sensing range to replicate a sensor like LiDAR. The noisy measurements were processed though Kalman filtering using the constant velocity model. The measurement noise is modeled as zero-mean Gaussian $v \sim \mathcal{N}(0, \mathcal{\bm  R})$ with variance of $\sigma_{\text{meas}} = 0.05$ m. 

\subsection{Single Vehicle Obstacle Avoidance}
In this section, the proposed method is evaluated with single quadcopter performing obstacle avoidance on moving and stationary obstacles. In the proposed scenario, a quadcopter equipped with the hydrodynamic-flow based collision avoidance algorithm flies in a figure-eight trajectory while encountering stationary (obstacle 1) and moving obstacles (obstacle 2). As a value for clearance size, $R_d$ value of 1 m was selected. 


\begin{figure}[hbt!]
	\centering
	\begin{subfigure}{0.45\textwidth}
		\centering
		\includegraphics[width = 1.0\textwidth]{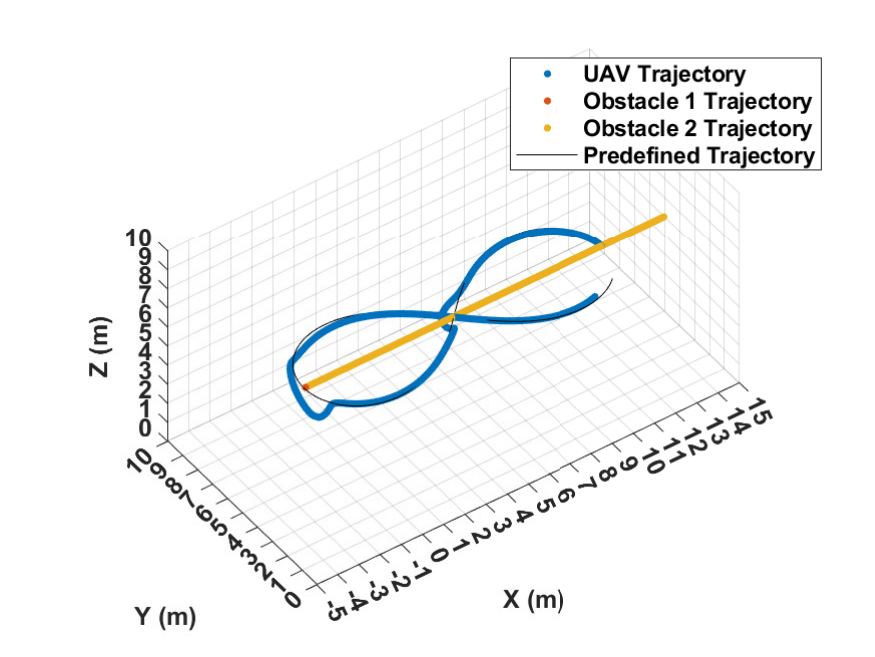}
		\caption{Quadcopter trajectory under hydrodynamic-based obstacle avoidance}
		\label{fig:single_uav_sim_trajectory}
	\end{subfigure}
	\begin{subfigure}{0.45\textwidth}
		\includegraphics[width = 1.0\textwidth]{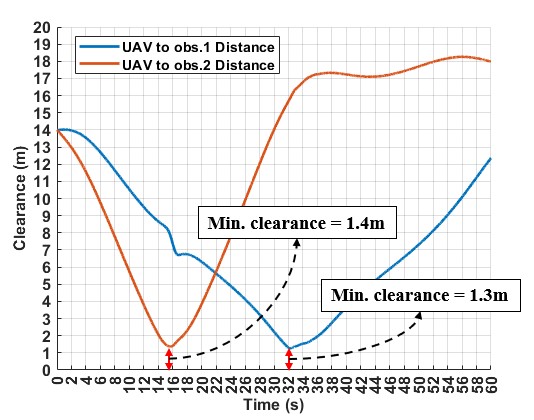}
		\caption{Clearance between the quadcopter and the obstacles}
		\label{fig:single_uav_sim_clearance}
	\end{subfigure}
	\caption{Single quadcopter simulation against stationary and moving obstacles with hydrodynamic-based obstacle avoidance}
	\label{fig:single_uav_sim}
\end{figure}

As Fig.~\ref{fig:single_uav_sim_trajectory} shows, avoidance of both stationary and moving obstacles are achieved in smooth manner with a sufficient clearance between the vehicle and the obstacles as shown in Fig.~\ref{fig:single_uav_sim_clearance}. In a case where the clearance need to be increased, the avoidance clearance can be adjusted with the value of $R_d$ shown in Eq.~(\ref{eq:streangth}) and the buffer radius $\epsilon$. 


\begin{figure}[hbt!]
\centering
\includegraphics[width = 1.0\textwidth]{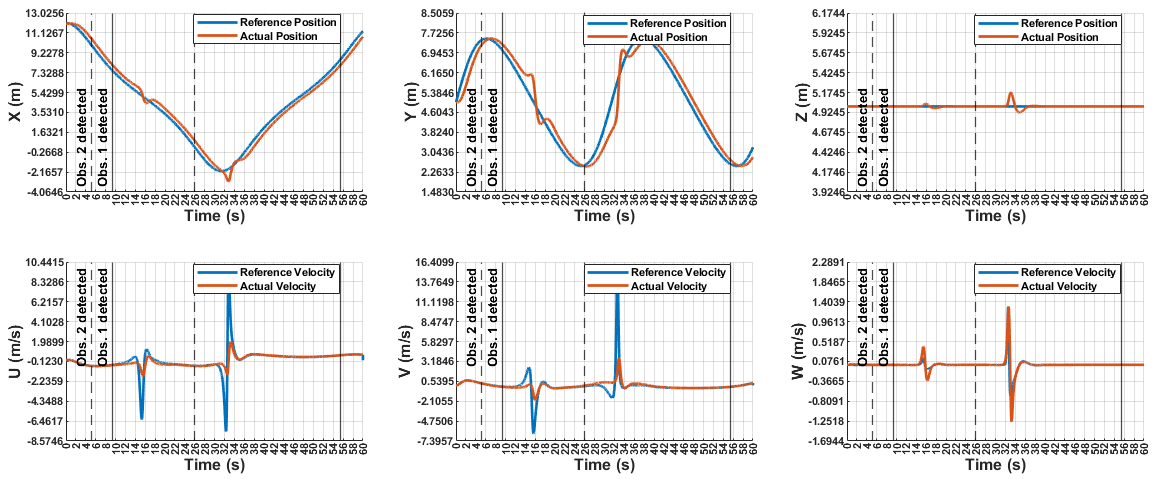}
\caption{Comparisons between reference states and UAV's actual states}
\label{fig:single_uav_ref_vs_true}
\end{figure}

Figure~\ref{fig:single_uav_ref_vs_true} shows comparisons between reference signal and actual UAV's response. As it shows, the reference trajectory is well tracked with larger deviation only when avoidance maneuvers are taken place. Since the vehicle constraints such as maximum RPM and employment of a simple PID controller, there are large errors between the reference velocity generated from the hydrodynamic-based collision avoidance algorithm and the actual vehicle's response. This result may be improved by utilizing optimal control method such as model predictive control (MPC) in our future study. More importantly, successful obstacle avoidance was accomplished under noisy data, showcasing the robustness of the method against noisy perception data. This result is particularly useful when considering the practical deployment of the algorithm in the future.

\subsection{Virtual Rigid Body (VRB) Formation}
Before simulating the synthesized model of VRB framework and the hydrodynamic flow-based collision avoidance, the performance of VRB formation algorithm is briefly introduced and evaluated in this section. 

\begin{figure}[hbt!]
\centering
\includegraphics[width = 0.65\textwidth]{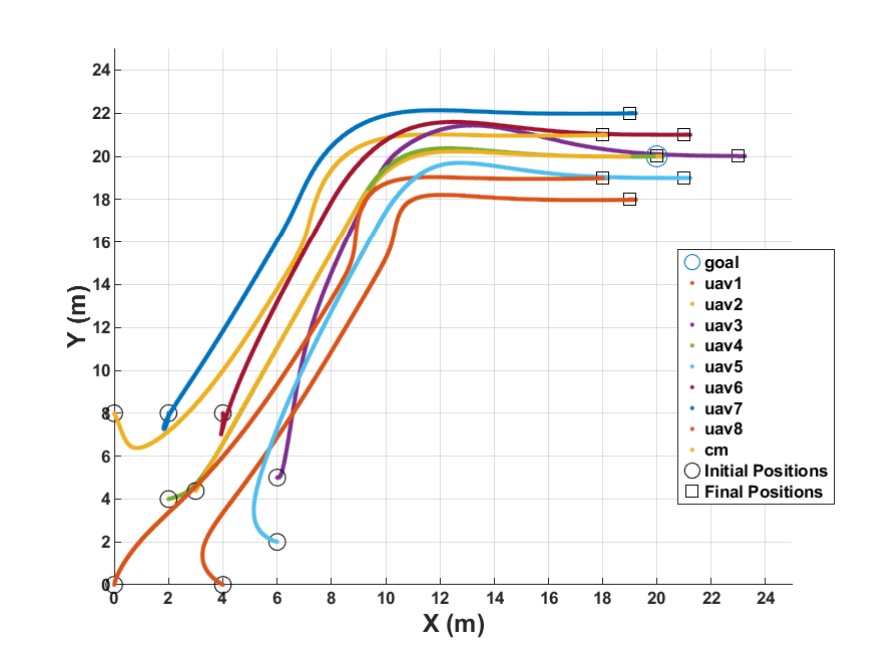}
\caption{UAV trajectories for VRB delta formation (Top View)}
\label{fig:vrb_sim}
\end{figure}

\begin{figure}[htb!]
\centering
\includegraphics[width=0.65\textwidth]{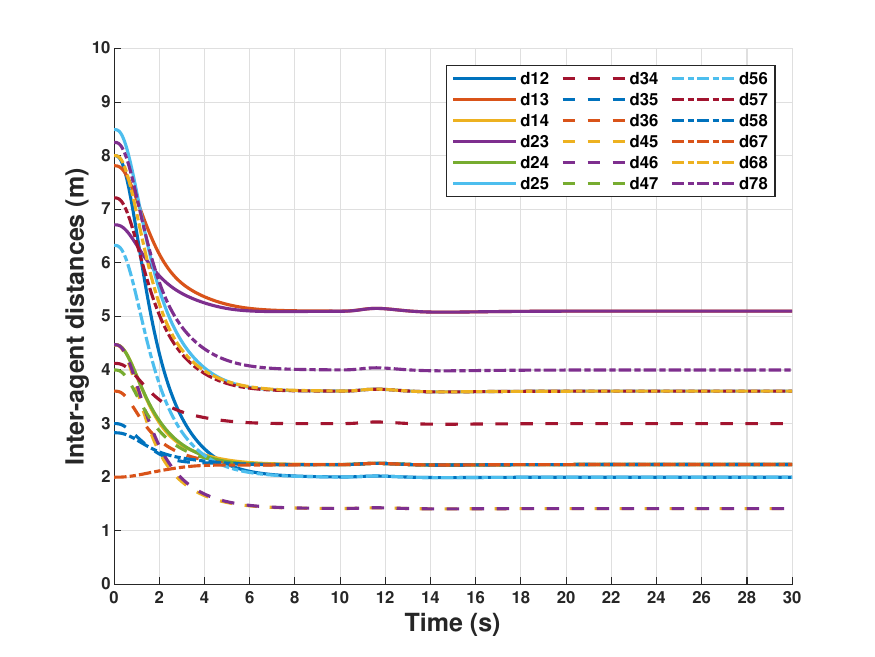}
\caption{history of Inter-agent distances}
\label{fig:vrb_sim_distances}
\end{figure}

The result in Fig.~\ref{fig:vrb_sim} shows eight vehicles engaging in a cooperative motion to establish and maintain a delta-shape formation. The desired formation is establish smoothly from arbitrary initial positions of the vehicles and maintained throughout the simulation. Moreover, since there are 8 vehicles, 18 constraints (i.e. $3 \cdot 8 - 6 = 18$) are required to guarantee rigidity of the formation. For the 18 constraints, vehicle pairs of $(v_i, v_j) = \{(v_1,v_2)$, $(v_1,v_3)$, $(v_1,v_4)$, $(v_2,v_3)$, $(v_2,v_4)$, $(v_2,v_5)$, $(v_3,v_4)$, $(v_3,v_5)$, $(v_3,v_6)$, $(v_4,v_5)$, $(v_4,v_6)$, $(v_4,v_7)$, $(v_5,v_6)$, $(v_5,v_7)$, $(v_5,v_8)$, $(v_6,v_7)$, $(v_6,v_8)$, $(v_7,v_8)\}$ are considered. These desired 18 inter-agent distances to achieve rigidness of the formation were successfully realized and stabilized although slight deviation was recorded when the orientation took place as shown in Fig.~\ref{fig:vrb_sim_distances}.

\begin{figure}[htb!]
\centering
\includegraphics[width = 1.0\textwidth]{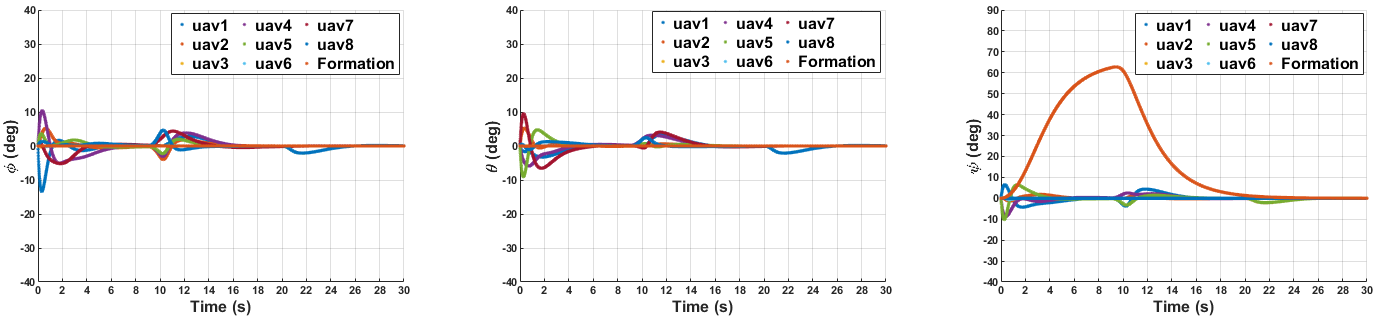}
\caption{UAV and VRB formation orientation profiles}
\label{fig:vrb_sim_8agent_v_orientation}
\end{figure}

Additionally, attitudes of UAVs as well as the rigid body formation itself is evaluated to validate the VRB framework is able to establish, maintain, and re-orient the formation without abrupt and impractical behaviors. As subfigures in Fig.~\ref{fig:vrb_sim_8agent_v_orientation} show, the formation establishment, maintenance, and re-orientation were accomplished while the system maintained nominal stability, which confirms the baseline robustness of the formation framework.

This formation is also considered in the evaluation of synthesized approach to see the effect of hydrodynamic flow-based collision avoidance on the VRB formation. Animation of this simulation is given in the previously provided link. To further reinforce the results presented with multiple cases, simulation videos of other formation shapes and number of vehicles are given in the following link: \url{https://www.youtube.com/playlist?list=PLX7ZK_42SnIBDlIFMG7a-u_iIPPMuSxSL}.

\subsection{VRB - Hydrodynamic-Based Obstacle Avoidance}

\begin{figure}[hbt!]
	\centering
	\begin{subfigure}{0.45\textwidth}
		\centering
		\includegraphics[width = 1.0\textwidth]{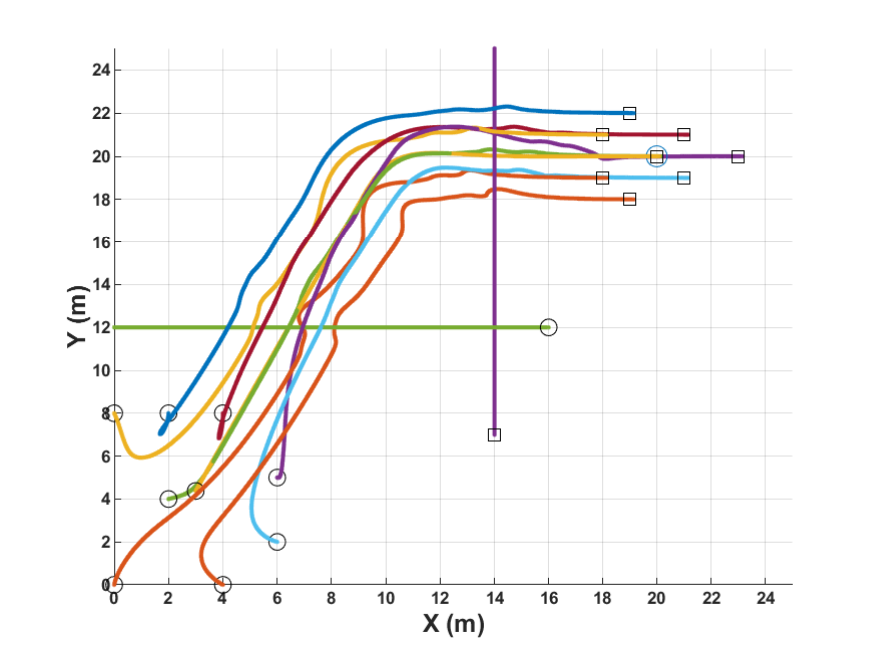}
		\caption{vehicle trajectories, top view}
	\end{subfigure}
	\begin{subfigure}{0.45\textwidth}
		\centering
		\includegraphics[width = 1.0\textwidth]{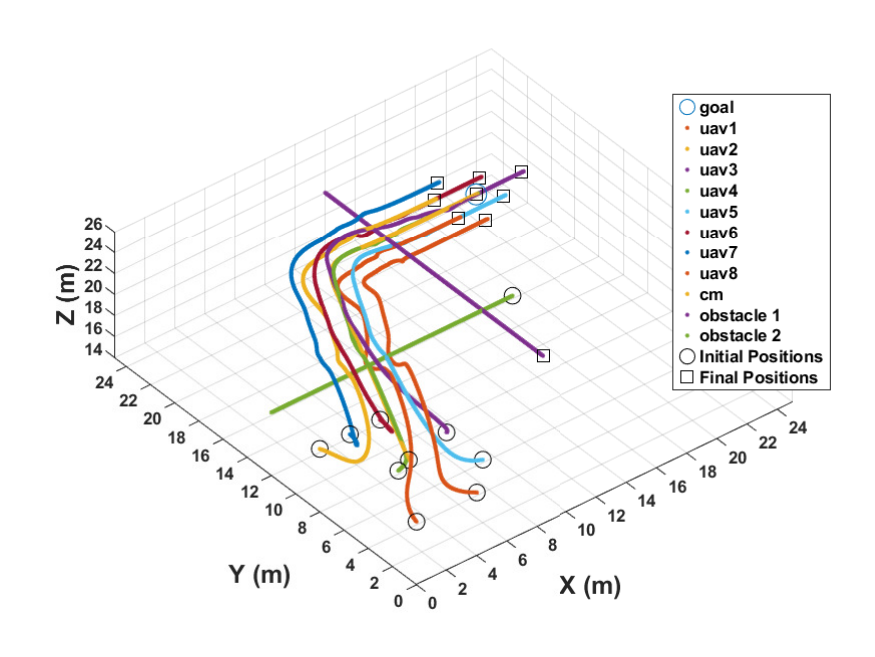}
		\caption{vehicle trajectories, isometric view}
	\end{subfigure}
	\vskip\baselineskip
	\begin{subfigure}{0.45\textwidth}
		\centering
		\includegraphics[width = 1.0\textwidth]{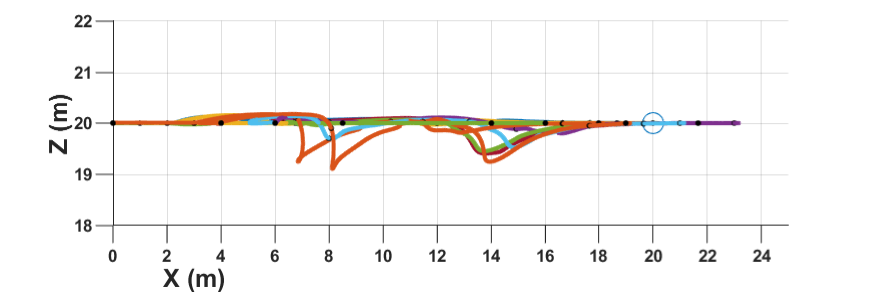}
		\caption{vehicle trajectory, front view}
	\end{subfigure}
	\begin{subfigure}{0.45\textwidth}
		\centering
		\includegraphics[width = 1.0\textwidth]{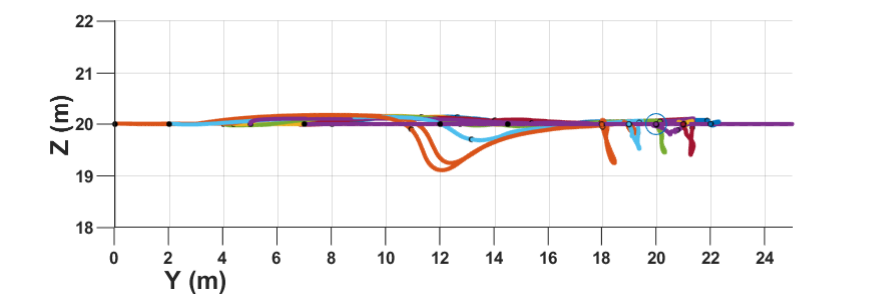}
		\caption{vehicle trajectory, side view}
	\end{subfigure}
	\caption{Eight-agent VRB formation in a Delta shape coordination with embedded hydrodynamic-based collision avoidance algorithm}
	\label{fig:hydro_vrb_sim}
\end{figure}

Building on the successful outcomes of the single vehicle hydrodynamic-based collision avoidance and multi-agent VRB formation, the proposed synthesized method is evaluated in a simulation. In the simulation, the same 8-agent delta-shape VRB formation is considered to compare the results obtained in the previous section and ones under the synthesized approach from this section. Gaussian noise is also applied to obstacles' positions to replicate sensor measurement, which is processed by the Kalman filter. By employing the algorithm shown in Algorithm~\ref{alg:algorithm}, the results shown in Fig.~\ref{fig:hydro_vrb_sim} were obtained. For the simulation settings, parameter values shown in Table~\ref{tab:hydro_vrb_sim_properties} are considered.

\begin{table}[hbt!]
\caption{VRB-Hydrodynamic-based Obstacle Avoidance Scenario Simulation Parameters}
\centering
\begin{tabular}{lcccccc}
\hline
\textbf{Parameter} & \textbf{Value}
\\\hline
Obstacle 1 initial position, ${\bm r_0}_{1,obs}$ & $[14,~37,~20]^T~$ m \\
Obstacle 2 initial position, ${\bm r_0}_{2,obs}$ & $[16,~12,~20]^T~$m \\
Obstacle 1 velocity, ${\bm v}_{1,obs}$ & $[0,~-0.75,~0]^T~$ m/s \\
Obstacle 2 velocity, ${\bm v}_{2,obs}$ & $[-0.75,~0,~0]^T~$ m/s \\
Desired clearance (buffer) size, $R_d$, ($\epsilon$) & $1$ m, ($1$ m) \\
Velocity tracking gain, $K_h$ & $0.3$ \\
Constraint force gain, $\alpha$, $\beta$, $\gamma$ & $\alpha=\beta=1$, $\gamma=0$ \\
\hline
\label{tab:hydro_vrb_sim_properties}
\end{tabular}
\end{table} 

As it is seen from Fig.~\ref{fig:hydro_vrb_sim}, smooth and continuous avoidance maneuvers were performed as the vehicles encountered the obstacles. To better visualize this result, a video of this simulation can be found in 1:32 \textasciitilde ~ of the previously provided supplemental video. Additionally, simulation videos of other formation geometries and number of vehicles embedded with this synthesized approach are given in the following link: \url{https://www.youtube.com/playlist?list=PLX7ZK_42SnIBaYop_a6Q5GKgtBj6kfUKp} to validate the performance of synthesized framework with different scenarios. 

\begin{figure}[htb!]
\centering
\begin{subfigure}{0.45\textwidth}
\includegraphics[width = 1.0\textwidth]{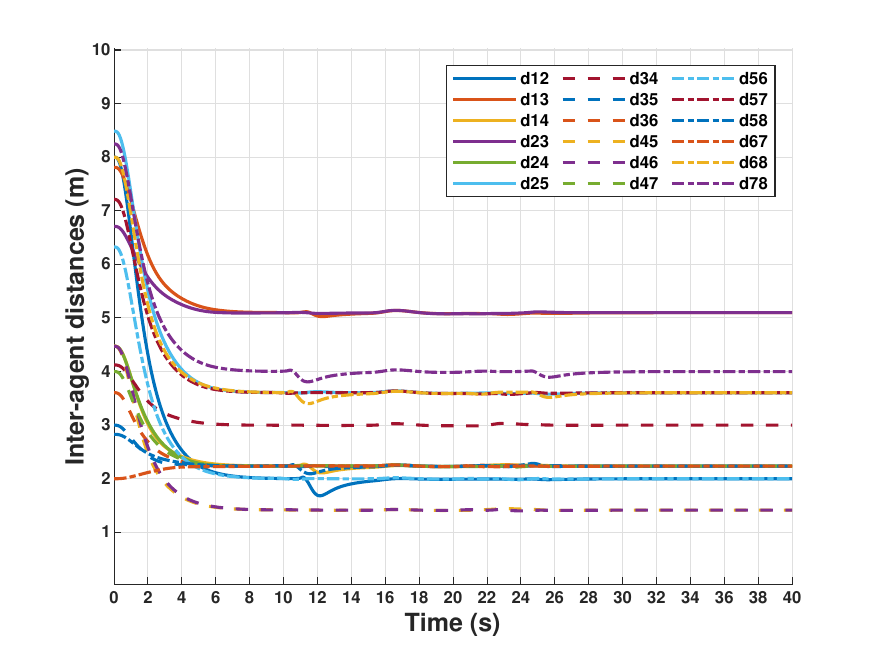}
\caption{Inter-agent distances for the 8-agent VRB formation with the hydrodynamic-based collision avoidance}
\label{fig:hydro_vrb_inter_agent_distances}
\end{subfigure}
\hspace{5mm}
\begin{subfigure}{0.45\textwidth}
\centering
\includegraphics[width = 1.0\textwidth]{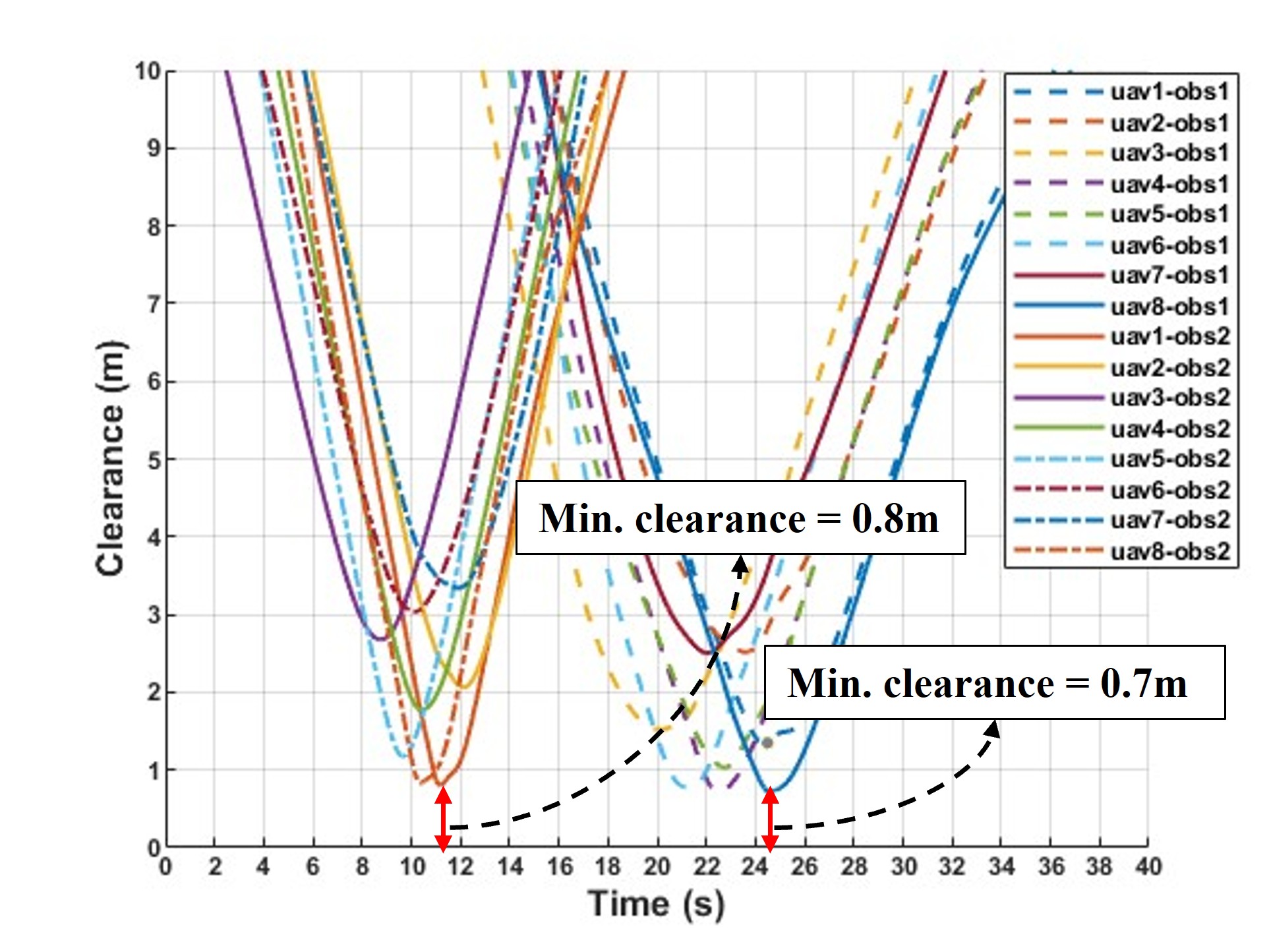}
\caption{Clearance between quadcopters and the obstacles}
\label{fig:hydro_vrb_clearance}
\end{subfigure}
\caption{Inter-agent distances and UAV-to-obstacle clearance}
\label{fig:hydro_vrb_distance_clearance}
\end{figure}

In addition to the results in Fig.~\ref{fig:hydro_vrb_sim}, inter-agent distances as well as profile of vehicle-to-obstacle clearances are provided in Fig.~\ref{fig:hydro_vrb_inter_agent_distances} and Fig.~\ref{fig:hydro_vrb_clearance}. As Fig.~\ref{fig:hydro_vrb_inter_agent_distances} shows, deviations in the inter-agent distances happen when the vehicles are performing avoidance maneuver against the obstacles. However, the desired values are achieved asymptotically once the obstacles are avoided thanks to the stable constraint force. Moreover, as it is evident from the supplemental video, when a vehicle is under an avoidance maneuver, other vehicles maneuver as well, trying to maintain the desired distances between them due to, again, the constraint force. This result is useful since it shows that the local collision avoidance between vehicles within the formation is already achieved by rigidness of the formation; therefore, the need for the consideration of collision avoidance algorithm within the formation is eliminated. 

\begin{figure}[htb!]
\centering
\includegraphics[width = 1.0\textwidth]{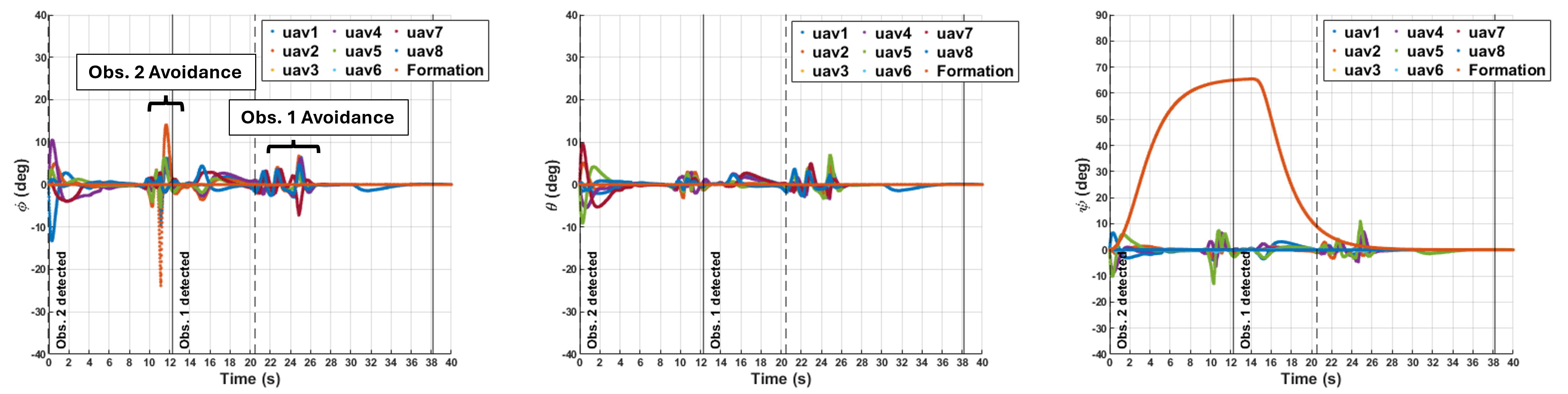}
\caption{UAV and VRB formation orientation profile within the collision avoidance scenario}
\label{fig:hydro_vrb_orientation}
\end{figure}

As it can be seen from Fig.~\ref{fig:hydro_vrb_clearance} and the supplemental video, both moving obstacles were avoided successfully although the clearance became less compared to the single vehicle scenario shown in Fig.~\ref{fig:single_uav_sim_clearance} since the vehicles are under influence of the constraint forces as well as the slot allocation control input. It is also seen from the supplemental video that the measurement noise affects the avoidance maneuver slightly as the vehicles oscillate as they avoid the obstacles. Lastly, attitudes of the vehicles and the formation, and comparison between the reference signals and vehicles' actual behaviors are provided in Fig.~\ref{fig:hydro_vrb_orientation}.

By comparing the figures in Fig.~\ref{fig:hydro_vrb_orientation} with those in Fig.~\ref{fig:vrb_sim_8agent_v_orientation}, orientation transients occur more actively and frequently in the avoidance scenario, corresponding to local attitude adjustments required for collision avoidance. It can be also interpreted from the figures and the supplemental video that frequent attitudes adjustments of not only vehicles executing the avoidance maneuver but also the surrounding vehicles are occurring due to the constraint force trying to maintain the formation while the avoidance maneuvers take place. Therefore, despite the individual deviations, the formation-level attitude remains stable, which indicates well-coordinated control among agents. Most importantly, the successful avoidance of moving obstacles under noisy sensor measurement data indicates that the proposed method is robust and could be considered as valid collision avoidance algorithm for both single vehicle scenario and formation flight scenario. In our future work, more robust control strategy such as MPC will be employed to leverage the potential of this framework.

\section{Conclusion}

Inspired by the natural behavior of fluid flow, this paper presented a three-dimensional hydrodynamic flow-based collision avoidance framework for both single and multiple UAVs in VRB formation operating in dynamic environments. By modeling moving obstacles as 3D doublets or ellipsoid that induce local velocity fields analogous to incompressible and irrotational fluid flow, UAVs were able to execute smooth, continuous, and collision-free avoidance maneuvers without explicit trajectory replanning or future state prediction. The integration of the VRB formation framework further ensured that the formation geometry and coordinated motion were maintained throughout avoidance maneuvers. Multiple simulation results validated the proposed algorithm could achieve smooth obstacle circumvention with sufficient clearance, even under noisy sensor data. The synthesized VRB-hydrodynamic approach enabled multiple UAVs to preserve formation rigidity and coordination while safely maneuvering around moving obstacles. The results also confirms that local collision avoidance within the formation is inherently achieved by the constraint forces of the VRB framework, eliminating the need for additional intra-formation avoidance strategies. Overall, the proposed approach offers a physics-inspired, computationally efficient, and scalable solution for real-time UAV collision avoidance and formation flight in cluttered and dynamic environments. Future work will focus on validation of the approach within higher fidelity simulation environment such as ROS2-Gazebo platform with software-in-the-loop (SITL) integration to incorporate realistic sensor data, flight control dynamics, and communication latency.

\newpage
\section*{Appendix}

\begin{algorithm}[htb!]
\caption{Pseudo Algorithm for VRB Formation with Hydrodynamic Collision Avoidance}
\begin{algorithmic}[1]

\Statex \textbf{--- Slot allocation of vehicles --- }
\If{$\pi^*$ is empty}
	\State$\pi^* \gets \arg \min_{\pi \in \Pi_N} \sum_{i=1}^N \| {\bm r}_i - {\bm S}_f(:, \pi(i)) \|_2$
	\State${\bm s} \gets {\bm S}_f(:, \pi^*) $
	\Comment{Assign optimal slot to each vehicle}
\EndIf

\For{$i = 1$ to $N$}

\State${\bm r}_{i,d} \gets {\bm r}_{f} + T_L^I {\bm s}_{i} $
\Comment{Reference position for $i$-th vehicle}

\State${\bm v}_{i,d} \gets {\bm v}_{f} + T_L^I ( {\bm \omega}_f \times {\bm s}_i )$
\Comment{Reference velocity for $i$-th vehicle}

\State${\bm f}_{i,s} \gets f({\bm r}_i, {\bm v}_i, {\bm r}_{i,d}, {\bm v}_{i,d})$
\Comment{Control input to fill in the assigned slot}

\EndFor

\State${\bm f}_s = [{\bm f}_{1,s}, {\bm f}_{2,s}, \ldots, {\bm f}_{N,s}]$

\Statex \textbf{--- Induced flow velocity due to obstacles --- }

\For{$i = 1$ to $N$}

\State${\bm v}_\infty \gets {\bm v}_i$
\Comment{Initialize the uniform flow}

\If{$N_{obs}$ obstacles are detected}

\For{$o = 1$ to $N_{obs}$}

\State$\hat{\bm r}_o,~\hat{\bm v}_o \gets KF({\bm r}_o)$
\Comment{Estimate the obstacle position and velocity} 

\State$r_{col} \gets \hat{\bm v}_{\infty,o} \cdot \hat{\bm r}_{i,o}~~\text{where}~~\hat{\bm v}_{\infty,o} = {\bm v}_\infty - \hat{\bm v}_{o},~\hat{\bm r}_{i,o} = {\bm r}_i - \hat{\bm r}_{o}$ 
\Comment{Assess collision risk}

\If{$r_{col} < 0$}
\Comment{If $o$-th obstacle is approaching ...}

\State $\{{\bm r}_{i,o}\}_D \gets {\bm T}_I^D (\theta_{EL}, \theta_{AZ}) \hat{\bm r}_{i,o}$
\State $\mu = -2 \pi (R_d + \epsilon)^3 \|\hat{\bm v}_{\infty,o}\|_2$
\Comment{Strength of the doublet with a buffer $\epsilon$}
\State $\{ {\bm v}_c \}_E \gets $ eq.(\ref{eq:combined_flow_local})
\Comment{Combined flow in the spherical coordinate system}
\State $\{ {\bm v}_c \}_I \gets {\bm T}_D^I {\bm T}_E^D(\theta, \Phi) \{ {\bm v}_c \}_E$
\Comment{Transformation to the Inertial frame}
\State ${\bm v}_\infty \gets \{ {\bm v}_c \}_I + \hat{\bm v}_o$
\Comment{Absolute uniform flow velocity}

\EndIf

\EndFor

\EndIf

\State${\bm v}_{i,h} \gets {\bm v}_\infty$
\Comment{Induced flow velocity on $i$-th vehicle due to the obstacles}
\State${\bm f}_h \gets f({\bm v}_i, {\bm v}_{i,h})$
\Comment{Control input to track the induced velocity}

\EndFor

\State${\bm f}_h \gets [{\bm f}_{1,h},~{\bm f}_{2,h},\ldots, {\bm f}_{N,h}]$

\Statex \textbf{--- Trajectory control input --- }
\For{$i = 1$ to $N$}

\State${\bm f}_{i,g} \gets [0,~0,~m_ig]^T$
\Comment{Force to compensate the gravitational force}
\State${\bm f}_{i,u} \gets {\bm f}_{i,h} + {\bm f}_{i,s} + {\bm f}_{i,g}$

\EndFor

\State${\bm f}_{u} \gets [{\bm f}_{1,u},~{\bm f}_{2,u},\ldots,{\bm f}_{N,u}]$

\Statex \textbf{--- Constraint force --- }
\State${\bm c} \gets$ eq.(\ref{eq:constraint_function}), $\dot{\bm c} \gets$ eq.(\ref{eq:velocity_constraint})
\State${\bm J} \gets \frac{\partial {\bm c}}{\partial {\bm r}}$
\State$\dot{\bm J} \gets \frac{d \bm J}{dt} = \frac{\partial {\bm J}}{\partial {\bm r}}\frac{d {\bm r}}{dt}$
\State${\bm f}_{e} \gets {\bm f}_u - {\bm f}_g$
\State${\bm f}_c \gets $ eq.(\ref{eq:constraint_force_comp})
\Comment{Constraint force}

\Statex \textbf{--- Total control input --- }
\State${\bm f}_{cmd} = {\bm f}_u + {\bm f}_c$
\Comment{Total control input for vehicles}

\State \texttt{Distribute the total control inputs to each vehicle}

\end{algorithmic}
\label{alg:algorithm}
\end{algorithm}

\bibliography{sample}

\end{document}